\documentclass[conference]{IEEEtran}

\IEEEoverridecommandlockouts                              

\usepackage{amsmath} 
\usepackage{amssymb}  
\usepackage{amsthm}
\usepackage{graphicx}
\usepackage{booktabs}

\usepackage{newfloat}
\usepackage{listings}
\usepackage{enumitem}
\usepackage{subcaption}
\usepackage{multirow}

\usepackage{algorithm}
\usepackage{algorithmic}
\usepackage{hyperref}

\usepackage[noadjust]{cite}

\usepackage{soul,xcolor}

\title{
Adaptive Querying for Reward Learning from Human Feedback
}

\author{Yashwanthi Anand, Nnamdi Nwagwu, Kevin Sabbe, Naomi T. Fitter and Sandhya Saisubramanian
}

\author{
\IEEEauthorblockN{
Yashwanthi Anand,
Nnamdi Nwagwu,
Kevin Sabbe,
Naomi T. Fitter,
Sandhya Saisubramanian
}
\IEEEauthorblockA{
Oregon State University\\
\{anandy, nwagwun, sabbek, naomi.fitter, sandhya.sai\}@oregonstate.edu
}
}

\begin{document}

\maketitle
\thispagestyle{empty}
\pagestyle{empty}

\begin{abstract}
Learning from human feedback is a popular approach to train robots to adapt to user preferences and improve safety. Existing approaches typically consider a single querying (interaction) format when seeking human feedback and do not leverage multiple modes of user interaction with a robot.
We examine how to learn a penalty function associated with unsafe behaviors using \emph{multiple} forms of human feedback, by optimizing both the \emph{query state} and \emph{feedback format}.
Our proposed \textit{adaptive feedback selection} is an iterative, two-phase approach which first selects critical states for querying, and then uses information gain to select a feedback format for querying across the sampled critical states. The feedback format selection also accounts for the cost and probability of receiving feedback in a certain format.
Our experiments in simulation demonstrate the sample efficiency of our approach in learning to avoid undesirable behaviors. The results of our user study with a physical robot highlight the practicality and effectiveness of adaptive feedback selection in seeking informative, user-aligned feedback that accelerate learning. Experiment videos, code and appendices are found on our website: \href{https://tinyurl.com/AFS-learning}{https://tinyurl.com/AFS-learning}

\end{abstract}

\section{INTRODUCTION}

\begin{figure}[t]
    \centering
    \includegraphics[width=\linewidth]{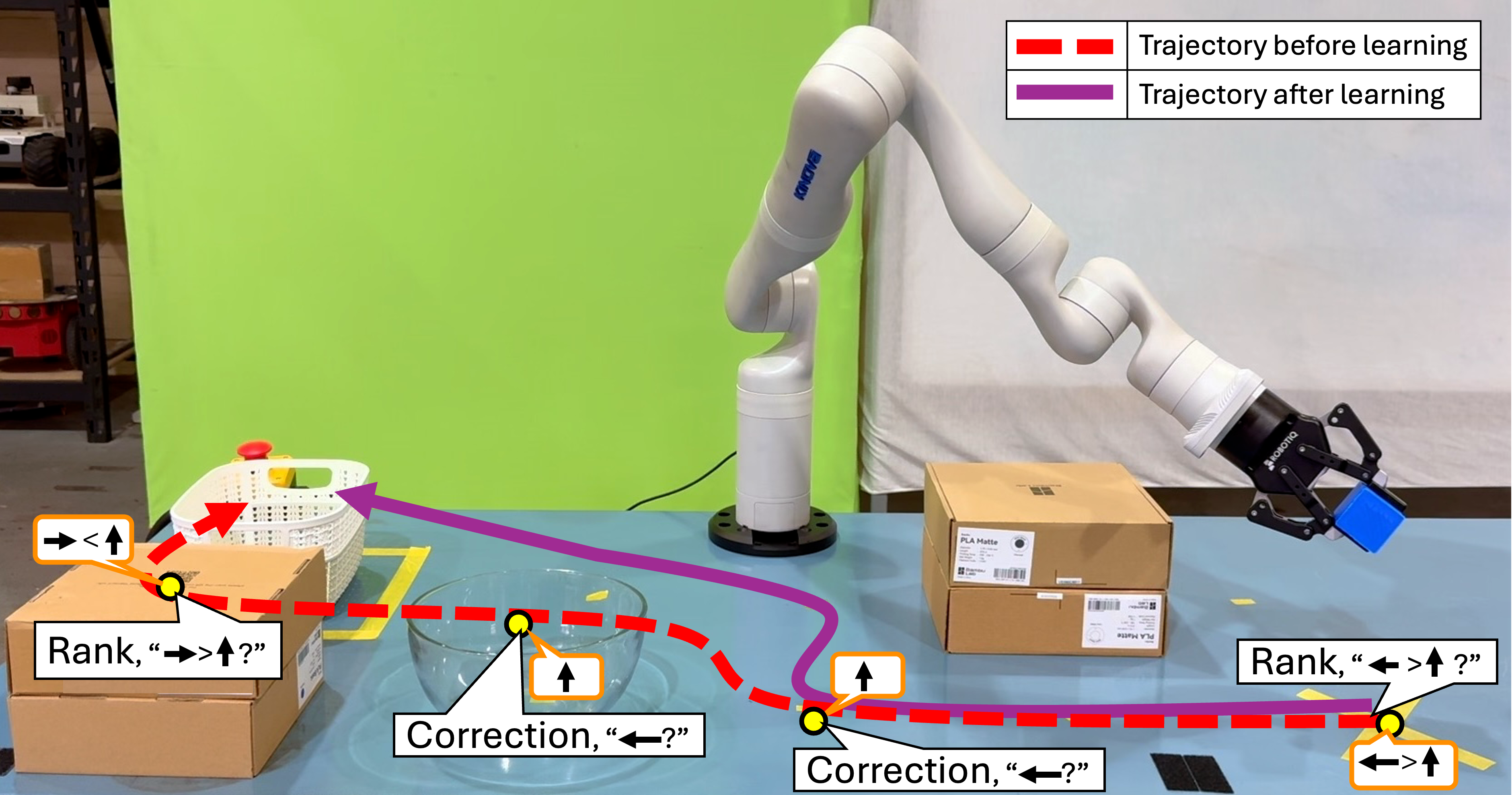}
    \caption{An illustration of adaptive feedback selection. The robot arm learns to move the blue object to the white bin, without colliding with other objects in the way, by querying the human in different format across the state space.}
    \label{fig:query}
\end{figure}

A key factor affecting an autonomous agent’s behavior is its reward function. Due to the complexity of real-world environments and the practical challenges in reward design, agents often operate with incomplete reward functions corresponding to underspecified objectives, which can lead to unintended and undesirable behaviors such as negative side effects (NSEs)~\cite{amodei2016concrete,saisubramanian2021multi,srivastava2023planning}.
For example, a robot optimizing the distance to transport an object to the goal, may damage items along the way if its reward function does not model the undesirability of colliding into other objects in the way (Figure~\ref{fig:query}).

Human feedback offers a natural way to provide the missing knowledge, and several prior works have examined learning from various forms of human feedback to improve robot performance, including avoiding side effects~\cite{cui2018active,cui2021empathic,lakkaraju2017identifying,ng2000algorithms,saran2021efficiently,zhang2020querying}.
In many real-world settings, the human can provide feedback in many forms, ranging from binary signals indicating action approval to correcting robot actions, each varying in the granularity of information revealed to the robot and the human effort required to provide it. For instance, a person supervising a household robot may occasionally be willing to provide detailed corrections when the robot encounters a fragile vase but may only want to give quick binary approvals during a routine motion. Ignoring this variability either limits what the robot can learn or burdens the user. To efficiently balance the \emph{trade-off} between seeking feedback in a format that accelerates robot learning and reducing human effort involved,
it is beneficial to seek detailed feedback sparingly in certain states and complement it with feedback types that require less human effort in other states. Such an approach could also reduce the sampling biases associated with learning from any one format, thereby improving learning performance~\cite{saisubramanian2022avoiding}.
In fact, a recent study indicates that users are generally willing to engage with the robot in more than one feedback format~\cite{saisubramanian2021understanding}. 
However, existing approaches rarely exploit this flexibility,
and \emph{do not support} gathering feedback in different formats in different regions of the state space~\cite{cui2021understanding,settles1995active}.

These practical considerations motivate the core question of this paper: ``How can a robot identify \emph{when to query} and in \emph{what format}, while accounting for the cost and availability of different forms of feedback?"
We present a framework for \emph{adaptive feedback selection} (AFS) that enables a robot to seek feedback in multiple formats in its learning phase, such that its information gain is maximized. 
Rather than treating all states and feedback formats uniformly, AFS prioritizes human feedback in states where feedback is most valuable and chooses feedback types based on their expected cost and information gain. This design reduces user effort, accommodates different levels of feedback granularity, and focuses on states where learning contributes to safety.
In the interest of clarity, the rest of this paper grounds the discussion of AFS as an approach for robots to learn to avoid negative side effects (NSEs) of their actions. The NSEs refer to unintended and undesirable outcomes that arise as the agent performs its assigned task. In object delivery example in Figure~\ref{fig:query}, the robot may inadvertently collide with other objects on the table, producing NSEs. Focusing on NSEs provides a well-defined and measurable setting--quantified by the number of NSE occurrences--to evaluate how AFS improves an agent's learning efficiency and safety. However, note that AFS is a general technique that can be applied broadly to learn about various forms of undesirable behavior.

Minimizing NSEs using AFS involves four iterative steps (Figure~\ref{fig:algo_overview}):
(1) states  are partitioned into clusters, with a cluster weight proportional to the number of NSEs discovered in it;
(2) a set of critical states---states where human feedback is crucial for learning an association of state features and NSEs, i.e., a predictive model of NSE severity, is formed by sampling from each cluster based on its weight; (3) a feedback format that maximizes the information gain in critical states is identified, while accounting for the cost and uncertainty in receiving a feedback, using the human feedback preference model; and (4) update cluster weights and information gain, and sample a new set of critical states to learn about NSEs, until the querying budget expires. The learned NSE information is mapped to a penalty function and augmented to the robot's model to compute an NSE-minimizing policy to complete its task. 

We evaluate AFS in both simulation and using a user study where participants interact with a robot arm. 
First, we evaluate the approach in three simulated proof-of-concept settings with simulated human feedback. Second, we conduct a pilot study where 12 human participants interact with and provide feedback to the agent in a simulated gridworld domain. Finally, we evaluate using a Kinova Gen3 7DoF arm and 30 human participants.
Besides the performance and sample efficiency, our experiments also provide insights into how the querying process can influence user trust. Together, these complementary studies demonstrate both the practicality and effectiveness of AFS. 

\section{BACKGROUND and RELATED WORK}
\subsection{Markov Decision Processes (MDPs)} The MDPs are a popular framework to model sequential decision making problems. An MDP is defined by the tuple $M\!=\!\langle S,A,T,R,\gamma \rangle$, where $S$ is the set of states, $A$ is the set of actions,
$T(s, a, s')$ is the probability of reaching state $s'\!\in\!S$ after taking an action $a\!\in\!A$ from a state $s\!\in\!S$ and $R(s, a)$ is the reward for taking action $a$ in state $s$.
An optimal deterministic policy $\pi^*:\!S\!\rightarrow\!A$ is one that maximizes the expected reward. 
When the objective or reward function is incomplete, even an optimal policy can produce unsafe behaviors such as side effects. \textbf{Negative Side Effects} (NSEs) are immediate, undesired, unmodeled effects of an agent's actions on the environment~\cite{DBLP:journals/corr/abs180601186,DBLP:conf/atal/SaisubramanianZ21,srivastava2023planning}.
We focus on NSEs arising due to incomplete reward function~\cite{saisubramanian2021multi}, which we mitigate by learning a penalty function using human feedback.

\subsection{Learning from Human Feedback}
Learning from human feedback is a popular approach to train agents when reward functions are unavailable or incomplete~\cite{abbeel2004apprenticeship,ng2000algorithms,ross2011reduction,10.3389/frobt.2021.584075}, including to improve safety~\cite{brown2020bayesian,pmlrv87brown18a,hadfield2017inverse,ramakrishnan2020blind,zhang2020querying,saisubramanian2021multi,hassan2025coherence}.
Feedback can take various forms such as \emph{demonstrations}~\cite{10.5555/1625275.1625692,saisubramanian2021multi,seo2024idil,zha2024learning}, \emph{corrections}~\cite{cui2023lilac,barmann2024incremental}, \emph{critiques}~\cite{cui2018active,10803055}, \emph{ranking} trajectories~\cite{brown2020safe,ijcai2024p586,feng2025duo}, natural language instructions~\cite{lou2024safe,yang2024multimodal,hassan2025coherence}
or may be \emph{implicit} in the form of facial expressions and gestures~\cite{cui2021empathic,10.3389/frobt.2022.838059,candon2023nonverbal}.

While the existing approaches for learning from feedback have shown success, they typically assume that a single feedback type is used to teach the agent. This assumption limits learning efficiency and adaptability. Some efforts combine demonstrations with preferences~\cite{biyik2022learning,ibarz2018reward}, showing that utilizing more than one format accelerates learning. Extending this idea, recent works integrate richer modalities such as language and vision with demonstrations. ~\cite{yang2024trajectory} learn reward function from comparative language feedback, while ~\cite{sontakke2023roboclip} show that a single demonstration or natural language description can help define a proxy reward when used along with a vision-language models (VLM) that is pretrained on a large amount of out-of-domain video demonstrations and language pairs. ~\cite{kim2023guide} use multimodal embeddings of visual observations and natural language descriptions to compute alignment-based rewards. A recent study even emphasizes that combining multiple feedback modalities can further enhance learning outcomes~\cite{beierling2025power}.
Together, these works highlight that combining complementary feedback formats help advance reward learning beyond using a fixed feedback format. In contrast, our approach uses multiple forms of human feedback for learning. 

Other approaches that learn from human feedback focus on modeling variations in human behavior.~\cite{huang2024modeling} model the heterogeneous behaviors of human, capturing differences in feedback frequency, delay, strictness, and bias to improve the robustness during the learning process, as optimal behaviors vary across users. Along the same line, the reward learning approach proposed by~\cite{ghosal2023effect}, selects a single feedback format based on the user ability to provide feedback in that format, resulting in an interaction that is tailored to a user's skill level. 
Collectively, these works reveal a shift towards adaptive and user-aware querying mechanisms that improve reward inference and learning efficiency, motivating our approach to dynamically select both when to query and in what feedback format.

\section{PROBLEM FORMULATION}

\noindent \textbf{Setting:} Consider a robot operating in a discrete environment modeled as a Markov Decision Process (MDP), using its acquired model $M = \langle S,A,T,R_T \rangle$. The robot optimizes the completion of its assigned task, which is its primary objective described by reward $R_T$. A \emph{primary policy}, $\pi^M$, is an optimal policy for the robot's primary objective.

\vspace{2pt}
\noindent \textbf{Assumption 1.} Similar to~\cite{saisubramanian2021multi}, we assume that the agent's model $M$ has all the necessary information for the robot to successfully complete its assigned task but lacks other superfluous details that are unrelated to the task.

\begin{figure*}[t]
    \includegraphics[width=\textwidth]{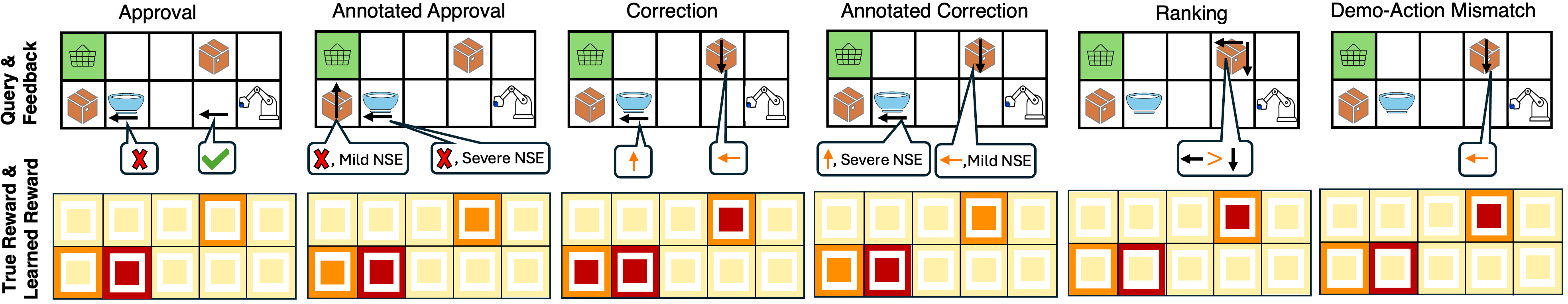}
    \caption{Visualization of reward learned using different feedback types. \textbf{(Row 1)} Black arrows indicate queries, and feedback is in speech bubbles. \textbf{(Row 2)} \raisebox{-1px}{\includegraphics[width=0.3cm]{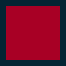}} denotes high, \raisebox{-1px}{\includegraphics[width=0.3cm]{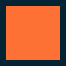}} mild, and \includegraphics[width=0.3cm]{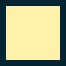} zero penalty. Outer box is the true reward, and inner box shows the learned reward. Mismatches between the outer and inner box colors indicate incorrect learned model.
    }
    \label{fig:reward_illustration}
\end{figure*}

\vspace{2pt}
Since the model is incomplete in ways unrelated to the primary objective, executing the primary policy produces negative side effects (NSEs) that are difficult to identify at design time. Following~\cite{saisubramanian2021multi}, we define NSEs as immediate, undesired, unmodeled effects of a robot's actions on the environment.
We focus on settings where the robot has \emph{no prior knowledge} about the NSEs of its actions or the underlying true NSE penalty function $R_N$. It learns to avoid NSEs by learning a penalty function $\hat{R}_N$ from human feedback that is consistent with $R_N$.

We target settings where the human can provide feedback in multiple ways and the robot can seek feedback in a \emph{specific} format such as approval or corrections.  This represents a significant shift from traditional active learning methods, which typically gather feedback only in a single format~\cite{ramakrishnan2020blind,saisubramanian2021multi,saran2021efficiently}. Using the learned $\hat{R}_N$,
the robot computes an NSE-minimizing policy to complete its task by optimizing: $R(s, a) = \theta_1 R_T(s, a) + \theta_2 \hat{R}_N(s, a),$
where
$\theta_1$ and $\theta_2$ are fixed, tunable weights denoting priority over objectives.

\noindent \textbf{Running Example:} We illustrate the problem using a simple object delivery task using a Kinova Gen3 7DoF arm shown in Figure~\ref{fig:query}. 
The robot optimizes delivering the blue block to the white bin, by taking the shortest path. However, passing through states with a cardboard box or a glass bowl constitutes an NSE. Since the robot has no prior knowledge about NSEs of its actions, it may inadvertently navigate through these states causing NSEs.

\vspace{2pt}
\noindent \textbf{Human's Feedback Preference Model:}
The feedback format selection must account for the cost and human preferences in providing feedback in a certain format.
The user's \textit{feedback preference model} is denoted by
$D = \langle \mathcal{F}, \psi, C\rangle$ where,
\begin{itemize}
    \item $\mathcal{F}$ is a predefined set of feedback formats the human can provide, such as demonstrations and corrections;
    \item $\psi: \mathcal{F} \rightarrow [0, 1]$ is the probability of receiving feedback in a format $f$, denoted as $\psi(f)$; and
    \item $C: \mathcal{F} \rightarrow \mathbb{R}$ is a cost function that assigns a cost to each feedback format $f$, representing the human's time or cognitive effort required to provide that feedback.
\end{itemize}

This work assumes the robot has access to the user's feedback preference model $D$---either handcrafted by an expert or learned from user interactions prior to robot querying, as in our user study experiments.
Abstracting user feedback preferences into probabilities and costs enables generalizing the preferences across similar tasks. We take the pragmatic stance that $\psi$ is independent of time and state, denoting the user's preference about a format, such as not preferring formats that require constant supervision of robot performance. While this can be relaxed and the approach can be extended to account for state-dependent preferences, obtaining an accurate state-dependent $\psi$ could be challenging in practice.

\vspace{2pt}
\noindent \textbf{Assumption 2.} Human feedback is immediate and accurate, when available.

Below, we describe the various feedback formats considered in this paper, and how the data from these formats are mapped to NSE severity labels. 

\subsection{Feedback Formats Studied}
\label{subsec:feedbackFormats}
The agent learns an association between state-action pairs and NSE severity, based on the human feedback provided in response to agent queries. The NSE categories we consider in this work are $\{\text{No NSE}, \text{Mild NSE}, \text{Severe NSE} \}$.
We focus on the following commonly used feedback types, each differing in the level of information conveyed to the agent and the human effort required to provide them.

\noindent \textbf{Approval (App):~} The robot randomly selects $N$ state-action pairs from all possible actions in critical states and queries the human for approval or disapproval. Approved actions are labeled as acceptable, while disapproved actions are labeled as unacceptable.

\noindent \textbf{Annotated Approval (Ann. App):~} An extension of Approval, where the human specifies the \textit{NSE severity} (or category) for each disapproved action in the critical states.

\noindent\textbf{Corrections (Corr):~}
The robot performs a trajectory of its primary policy in the critical states, under human supervision. If the robot's action is unacceptable, then the human intervenes with an acceptable action in these states. If all actions in a state lead to NSE, the human specifies an action with the least NSE. When interrupted, the robot assumes all actions except the correction are unacceptable in that state.

\noindent \textbf{Annotated Corrections (Ann. Corr):~} An extension of Corrections, where the human specifies the severity of NSEs caused by the robot's unacceptable action in critical states.

\noindent \textbf{Rank:~} The robot randomly selects $N$ ranking queries of the form $\langle \textit{state}, \textit{action 1}, \textit{action 2}\rangle$, by sampling two actions for each critical state. The human selects the safer action among the two options. If both are safe or unsafe, one of them is selected at random. The selected action is marked as acceptable and the other is treated as unacceptable.

\noindent\textbf{Demo-Action Mismatch (DAM):~}
The human demonstrates a safe action in each critical state, which the robot compares with its policy. All mismatched robot's actions are labeled as unacceptable. Matched actions are labeled as acceptable.

\noindent \textbf{Mapping feedback data to NSE severity labels:} We use $l_a$, $l_m$, and $l_h$ to denote labels corresponding to no, mild and severe NSEs, respectively. An acceptable action in a state is mapped to $l_a$, i.e., $(s, a) \rightarrow l_a$, while an unacceptable action is mapped to $l_h$. When the severity of NSEs for unacceptable actions is known, actions producing mild NSEs are mapped to $l_m$ and those producing severe NSEs to $l_h$. Mapping feedback to this common label set provides a consistent representation of NSE severity across diverse feedback types.
The granularity of information and the sampling biases of the different feedback types affect the learned reward. Figure~\ref{fig:reward_illustration} illustrates this with the learned NSE penalty for the running example of moving an object to the bin (Fig.~\ref{fig:query}), motivating the need for an adaptive approach that can learn from more than one feedback format. In the running example, the robot arm colliding with cardboard boxes is a mild NSE, and colliding with a glass bowl is a severe NSE.

\section{ADAPTIVE FEEDBACK SELECTION}\label{sec:AFS}
Given an agent's decision making model $M$ and the human's feedback preference model $D$, AFS enables the agent to query for feedback in critical states in a format that maximizes its information gain. We first formalize the NSE model learning process and then describe in detail how AFS selects critical states and the query format. 

\noindent \textbf{Formalizing NSE Model Learning:}\label{Sec:NSEModel}
Let $p^*: S \times A \rightarrow \{l_a, l_m, l_h\}$ denote the \emph{true} NSE severity label for each state-action pair, which is unknown to the agent but known to the human. The label $l_a$ corresponds to \textit{no NSE}, $l_m$ denotes \textit{mild NSE}, $l_h$ denote the label for \textit{severe NSE}. Let $p$ be a sampled approximation of $p^*$ ($p \sim p^*$), denoting the dataset of NSE labels collected via human feedback in response to the $(s,a)$ pairs queried. That is, $p^t$ denotes the data collected from human feedback until iteration $t$, where $p^t(s,a)$ represents the categorical NSE severity label assigned to the state-action pair $(s,a)$. Let $q: S \times A \rightarrow \{l_a, l_m, l_h\}$ denote the labels predicted by the learned NSE model---learned using a supervised classifier with $p$ as the training data. In this paper, we use a Random Forest (RF) classifier, though any classifier can be used in practice. Hyperparameters are optimized through randomized search with three-fold cross validation, and the configuration yielding the lowest mean-squared error is selected for training. 

\begin{figure}[ht]
    \centering
    \includegraphics[width=\linewidth]{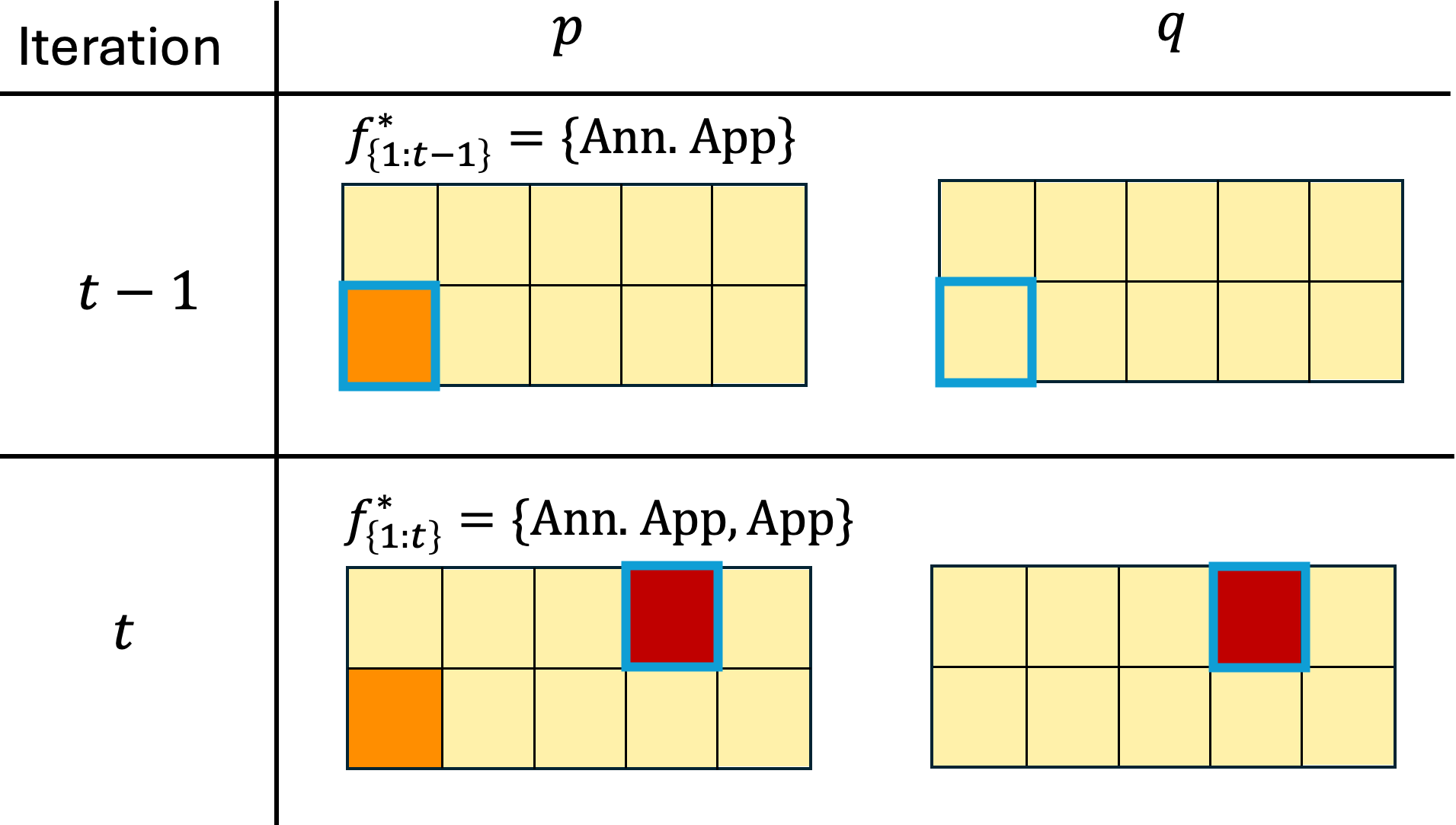}
    \caption{Illustration of $p$ (accumulated feedback) and $q$ (generalized NSE labels) for the object delivery task. $f^*_{1:t-1}$ indicates the feedback formats selected until iteration $t-1$. 
  \includegraphics[height=0.9em]{low_penalty.png} indicates no NSE;
  \includegraphics[height=0.9em]{mild_penalty.png} indicates mild NSE;
  \includegraphics[height=0.9em]{high_penalty.png} indicates severe NSE. Queried states in each iteration is highlighted in blue.}
    \label{fig:pq_labels}
\end{figure}

Figure~\ref{fig:pq_labels} shows an example of $p$ and $q$ for the object delivery task. We encode NSE categories as $\{0,1,2\}$ corresponding to $\{$ no NSE, mild NSE, severe NSE$\}$ respectively. Each state has four possible actions $A\!=\!\{a_1, a_2, a_3, a_4\}$, and the vector $p(s)\!=\!\{\cdot, \cdot, \cdot, \cdot\}$ (and similarly $q(s)$) encodes the categorical NSE labels for $(s,a_1), (s,a_2), (s,a_3), (s,a_4)$ in that order. Since the human's categorization of NSE is initially unknown, $p(s)$ is sampled from a uniform prior over the labels, and $q(s)$ is initialized to $[0,0,0,0]$ (all actions are assumed to be safe) across all states.

At $t-1$, $p^{t-1}$ reflects a single labeled state from the feedback received, while $q^{t-1}$ reflects NSE label for the state after learning from $p^{t-1}$. 
For example, in iteration $t\!-\!1$, an action $a_3$ in state $s$ is randomly selected for querying using the \textit{Annotated Approval} feedback format. The human labels it as mild NSE, so $p^{t-1}(s,a_3)=1$, and consequently $p^{t-1}(s)=[0, 0, 1, 0]$. 
After training on $p^{t-1}$, the classifier may sometimes incorrectly predict $q^{t-1}(s)\!=\![0,0,0,0]$, especially in early iterations when there is less data. At the next iteration $t$, the agent queries in a similar state using the \emph{Approval} format, where the action $a_1$ is randomly selected. Because the NSE severity level (i.e., mild/severe) cannot be indicated through the Approval format, $p^t$ is updated as $p^t(s)=[2,0,0,0]$, and training now yields a prediction $q^t(s)=[2,0,1,0]$ (i.e. the NSE model predicts severe NSE outcome on $a_1$ and a mild NSE outcome on $a_3$). This illustrates that $q$ may initially disagree with $p$, but as feedback accumulates on related states, the generalization of $q$ across actions begins to align with $p$.

Each predicted label is then mapped to a penalty value to form the learned penalty function, $\hat{R}_N$, with penalties for $l_a, l_m$ and $l_h$ set to $0, -5$ and $-10$ respectively, in our experiments. This penalty function is integrated into the agent's reward model to compute an updated policy that minimizes NSEs while completing the primary task.

\begin{figure*}[t]
\centering    
\includegraphics[width=0.92\textwidth]{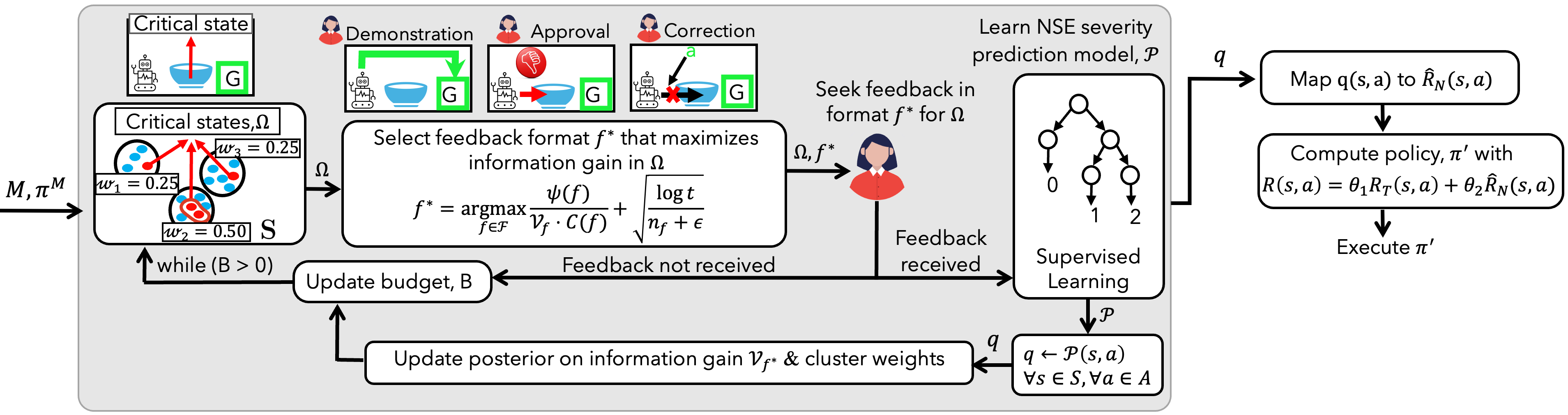}
    \caption{Solution approach overview. The critical states $\Omega$ for querying are selected by clustering the states. A feedback format $f^*$ that maximizes information gain is selected for querying the user across $\Omega$. The NSE model is iteratively refined based on feedback. An updated policy is calculated using a penalty function $\hat{R}_N$, derived from the learned NSE model.
    }
    \label{fig:algo_overview}
\end{figure*}

In this learning setup, minimizing NSEs using AFS involves four iterative steps (Figure~\ref{fig:algo_overview}). In each learning iteration, AFS identifies (1) which states are most critical for querying (Sec.~\ref{sec:criticalStates}), and (2) which feedback format maximizes the expected information gain at the critical states, while accounting for user feedback preferences and effort involved (Sec.~\ref{Sec:format_selection}). The information gain associated with a feedback quantifies the effect of a feedback in improving the agent's understanding of the underlying reward function, and is measured using Kullback-Leibler (KL) Divergence~\cite{ghosal2023effect,tien2023causal}. At the end of each iteration, the cluster weights and information gain are updated, and a new set of critical states are sampled to learn about NSEs, until the querying budget expires or the KL-divergence is below a problem-specific, pre-defined threshold.

\subsection{Critical States Selection}\label{sec:criticalStates}
When the budget for querying a human is limited, it is useful to query in states with a high \emph{learning gap} measured as the KL-divergence between the agent's knowledge of NSE severity and the true NSE severity given the feedback data collected so far. States with a high learning gap are called \emph{critical states} ($\Omega$) and querying in these states can reduce the learning gap.

\begin{algorithm}[t]
\caption{Critical States Selection}
\label{select_cs}
    \begin{algorithmic}[1]
        \REQUIRE $N$: \#critical states; $\mathcal{K}$:\#clusters; 
        \STATE $\Omega \leftarrow \emptyset$
        \STATE Cluster states into $\mathcal{K}$ clusters, $K = \{k_1, \ldots, k_{\mathcal{K}}\}$
        \FOR{each cluster $k \in K$}
            \STATE $
                W_k \leftarrow
                \begin{cases}
                    \frac{1}{\mathcal{K}}, \text{ if no feedback received in any iteration} \\
                    \frac{IG(k)}{\sum_{k \in K} IG(k)}, \text{ if feedback received}
                \end{cases}
            $
            \STATE $n_k \leftarrow \max(1, \lfloor W_k \cdot N \rfloor)$
            \STATE Sample $n_k$ states at random, $\Omega_k \leftarrow \text{Sample}(k, n_k)$
            \STATE $\Omega \leftarrow \Omega \cup \Omega_k$
        \ENDFOR
        \STATE $N_r \leftarrow N - |\Omega|$
        \IF{$N_r > 0$}
            \STATE $k' \leftarrow \arg\max_{k \in K} W_k$
            \STATE $\Omega \leftarrow \Omega \cup \text{Sample}(k', N_r)$
        \ENDIF
        \RETURN Set of selected critical states $\Omega$
    \end{algorithmic}
\end{algorithm}

Since $p^t$ and $q^t$ contain categorical values rather than probabilities, their corresponding empirical probability mass functions (PMFs) are computed over the three NSE categories (no NSE, mild NSE, and severe NSE), yielding $\hat{p}^t$ and $\hat{q}^{t}$, respectively. In this case, $\hat{p}^t$ and $\hat{q}^{t}$ will be vectors of length three, since we consider three NSE categories.

In order to select critical states for querying, we compute the KL divergence between  $\hat{q}^{t-1}$ and $\hat{p}^{t}$, $D_{KL}(\hat{p}^{t} \Vert \hat{q}^{t-1})$. Although $D_{KL}(\hat{p}^{t} \Vert \hat{q}^{t})$ may appear as a reasonable criterion to guide critical states selection, it only measures how well the agent learns from the feedback at $t$. It does not reveal states where the agent's predictions were incorrect.
For the example shown in Figure~\ref{fig:pq_labels} with $q^{t-1}(s)=[0,0,0,0]$ and $p^t(s)=[2,0,0,0]$, $\hat{p}^t$ and $\hat{q}^{t-1}$ are calculated as the average occurrence of each NSE category (no NSE, mild NSE, severe NSE) across the four actions. That is, for $q^{t-1}(s)=[0,0,0,0]$, the frequency is $[\frac{4}{4}, \frac{0}{4}, \frac{0}{4}]$, resulting in $\hat{q}^{t-1}(s)=[1.0, 0.0, 0.0]$. For $p^t(s)=[2,0,0,0]$, the frequency is $[\frac{3}{4}, \frac{0}{4}, \frac{1}{4}]$, yielding $\hat{p}^t(s)=[0.75, 0.0, 0.25]$. Calculating the divergence between $\hat{p}^t(s)$ and $\hat{q}^{t-1}(s)$ reveals that the prediction was incorrect at $s$ and therefore more data is required to align the learned model, and hence $s$ or similar states should be selected for querying. Therefore, the sampling weight of the cluster containing $s$ is increased (the region where the NSE model is still uncertain). In the following iteration, critical states are drawn from the reweighted clusters. Algorithm~\ref{select_cs} outlines our approach for selecting critical states at each learning iteration, with the following three key steps.

\noindent \emph{\underline{1. Clustering states}}:
Since NSEs are typically correlated with specific state features and do not occur at random, we cluster the states $S$ into $\mathcal{K}$ number of clusters so as to group states with similar NSE severity~\cite{lakkaraju2017identifying}.
In our experiments, we use KMeans clustering algorithm with Jaccard distance to measure the distance between states based on their features. In practice, any clustering algorithm can be used, including manual clustering. The goal is to create meaningful partitions of the state space to guide critical states selection for querying the user.

\noindent \emph{\underline{2. Estimating information gain}}: We define the information gain of sampling from a cluster $k\!\in\!K$, based on the learning gap, as follows:
\begin{align}
    IG(k)^t\!&=\!\frac{1}{|\Omega_{k}^{t-1}|} \sum_{s \in \Omega_{k}^{t-1}} D_{KL}(\hat{p}^{t}\Vert \hat{q}^{t-1}) \\
 \!&=\!\frac{1}{|\Omega_{k}^{t-1}|} \sum_{s \in \Omega_{k}^{t-1}} \sum_{l \in \{l_a, l_m, l_h\}} \hat{p}^{t}(l|s)\cdot \log \left(\frac{\hat{p}^{t}(l|s)}{\hat{q}^{t-1}(l|s)}\right),
 \label{Eqn: IG}
\end{align}
where $\Omega_k^{t-1}$ denotes the set of states sampled for querying from cluster $k$ at iteration $t-1$. $\hat{p}^t(l|s)$ and $\hat{q}^{t-1}(l|s)$ denote the probability of observing NSE category $l\in\{l_a, l_m, l_h\}$ in state $s$, derived from $p^t$ and $q^t$, respectively. This formulation quantifies how much the predicted NSE distribution diverges from the feedback received for each state, providing a principled measure of the expected information gain from querying in a cluster, $k$.

\noindent \emph{\underline{3. Sampling critical states}:} At each learning iteration $t$, the agent assigns a weight $w_k$ to each cluster $k\!\in\!K$, proportional to the new information on NSEs revealed by the most informative feedback format identified at $t-1$, using Eqn.~\ref{Eqn: IG}. Clusters are given equal weights when there is no prior feedback (Line 4).
Let $N$ denote the number of critical states to be sampled in every iteration. We sample critical states in batches but they can also be sampled sequentially. When sampling in batches of $N$ states, the number of states $n_k$ to be sampled from each cluster is determined by its assigned weight. At least one state is sampled from each cluster to ensure sufficient information for calculating the information gain for every cluster (Line 5).
The agent randomly samples $n_k$ states from corresponding cluster and adds them to a set of critical states $\Omega$ (Lines 6, 7).
If the total number of critical states sampled is less than $N$ due to rounding, then the remaining $N_r$ states are sampled from the cluster with the highest weight and added to $\Omega$ (Lines 9-12).

\begin{algorithm}[t]
\caption{Feedback Selection for NSE Learning}
\label{learn}
    \begin{algorithmic}[1]
        \REQUIRE B, Querying budget; $D$, Human preference model; $\delta$: KL divergence threshold

        \STATE $t \leftarrow 1$; $\mathcal{V}_f \leftarrow 0$ and $n_f\leftarrow 0,\ \forall f \in \mathcal{F}$
        \STATE Initialize $p$ and $q$: \hfill // $p$: random initialization, $q$: all safe
        \\$\forall s \in S, \forall a \in A,$ 
        $p(s,a) \leftarrow$ RandomNSELabel$(\{l_a, l_m, l_h\})$;
        $q(s,a) \leftarrow l_a$ 
        \WHILE {$B>0$ or $\forall s\in S, D_{KL}(\hat{p}^{t}\Vert \hat{q}^t)\leq\delta$}
            \STATE Sample critical states using Algorithm~\ref{select_cs}
            \STATE Query user with feedback format $f^*$, selected using using Eqn.~\ref{Eqn:select_format}, across sampled $\Omega$
            \IF {feedback received in format $f^*$ }
                \STATE $p^t \leftarrow$ Update distribution based on the feedback received in format $f^*$
                \STATE $\mathcal{P} \leftarrow$ TrainClassifier$(p^t)$
                \STATE $q^t \leftarrow \{ \mathcal{P}(s, a), \forall a \in A, \forall s \in \Omega\}$
                \STATE Update $\mathcal{V}_{f^*}$, using Eqn.~\ref{eqn:V_f-batch}
                \STATE $n_{f^*} \leftarrow n_{f^*}+ 1$
            \ENDIF
            \STATE $B \leftarrow B - C(f^*)$; $t \leftarrow t + 1$
        \ENDWHILE
        \RETURN NSE classifier model, $\mathcal{P}$
    \end{algorithmic}
\end{algorithm}

\subsection{Feedback Format Selection}
\label{Sec:format_selection}
To query in the critical states, $\Omega$, it is 
important to select a feedback format that not only maximizes the expected information gain about NSEs but also accounts for likelihood and  cost of the feedback. 
The \emph{information gain} of a feedback format $f$ at iteration $t$, for $N\!=\!|\Omega|$ critical states, is computed as the KL divergence between the observed and predicted NSE severity distributions, $\hat{p}^t$ and $\hat{q}^t$:
\begin{align}
    \mathcal{V}_f
    &=  \frac{1}{N} \sum_{s \in \Omega} D_{KL}(\hat{p}^{t}\Vert \hat{q}^t) \cdot \mathbb{I}[f\!=\!f_H^t] + \mathcal{V}_f \cdot (1-\mathbb{I}[f\!=\!f_H^t]),
    \label{eqn:V_f-batch}
\end{align}
where, $\mathbb{I}[f\!=\!f_H^t]$ is an indicator function that checks whether the format provided by the human, $f_H^t$, matches the requested format $f$. If no feedback is received, the information gain for that format remains unchanged.
The following equation is used to select the feedback format $f^*$, accounting for feedback cost and user preferences:
\begin{equation}
    f^* = \operatorname*{argmax}_{f \in \mathcal{F}} \underbrace{\frac{\psi(f)}{\mathcal{V}_f\cdot\text{C}(f)} + \sqrt{\frac{\log t}{n_f + \epsilon}}}_{\text{Feedback utility of $f$}},
    \label{Eqn:select_format}
\end{equation}
where $\psi(f)$ is the probability of receiving a feedback in format $f$ and $C(f)$ is the feedback cost, determined using the human preference model $D$. $t$ is the current learning iteration, $n_f$ is the number of times feedback in format $f$ was received, and $\epsilon$ is a small constant for numeric stability. The selected format $f^*$ represents the most informative feedback format given the agent's current knowledge, balancing exploration (less frequently used formats) and exploitation (formats known to provide high information gain).

\begin{figure*}[t]
    \centering
    \includegraphics[scale=0.6]{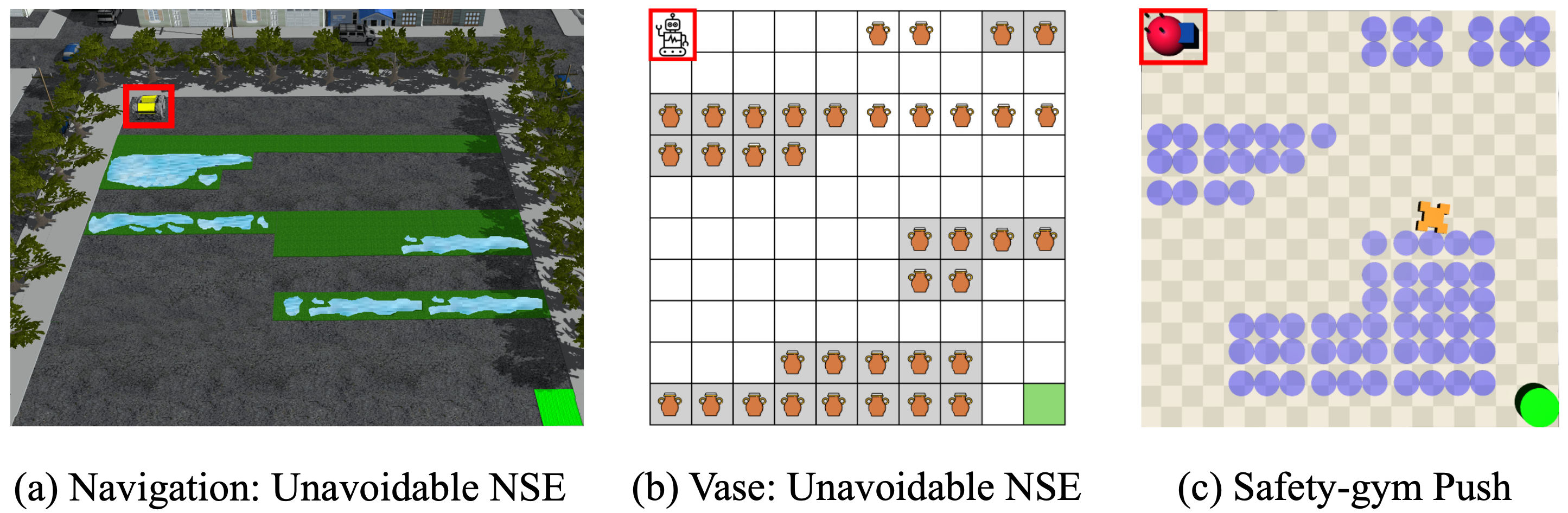}
    \caption{Illustrations of evaluation domains. Red box denotes the agent and the goal location is in green.}
\label{fig:SimDomains}
\end{figure*}

Algorithm~\ref{learn} outlines our feedback format selection approach. Since the agent has no prior knowledge of how the human categorizes NSE for each state-action pairs, the labeling function $p$ is instantiated by sampling from a uniform prior over the three NSE labels ($l_a, l_m, l_h$) for every $(s,a)$, while q is initialized assuming all actions are safe ($l_a$) (Line 2). These initial labels are progressively refined as human feedback is received. At each iteration, the agent samples $|\Omega|$ critical states using Algorithm~\ref{select_cs} (Line 4), and selects a feedback format $f^*$ is selected using Eqn.~\ref{Eqn:select_format}. The agent queries the human for feedback in $f^*$ (Line 5).
If the feedback is received (with probability $\psi(f^*)$), the observed NSE labels $p^t$ are updated and an NSE prediction model $\mathcal{P}$ is trained (Lines 6-8). The classifier $\mathcal{P}$ predicts the labels for the sampled critical states $\Omega$, yielding $q^t$. We restrict the prediction to $\Omega$ since these states indicate regions of high uncertainty and contribute to reducing the divergence between the true and learned NSE distributions. Further, restricting predictions to $\Omega$ also reduces computational overhead during iterative querying. $\mathcal{V}_{f^*}$ recomputed using Eqn.~\ref{eqn:V_f-batch}, and $n_{f^*}$ is incremented (Lines 9-11).
This repeats until either the querying budget is exhausted or the KL divergence between $\hat{p}^{t}$ and $\hat{q}^t$ over all states is within a problem-specific threshold $\delta$.  

\begin{figure}[t]
    \centering
        \includegraphics[width=\linewidth]{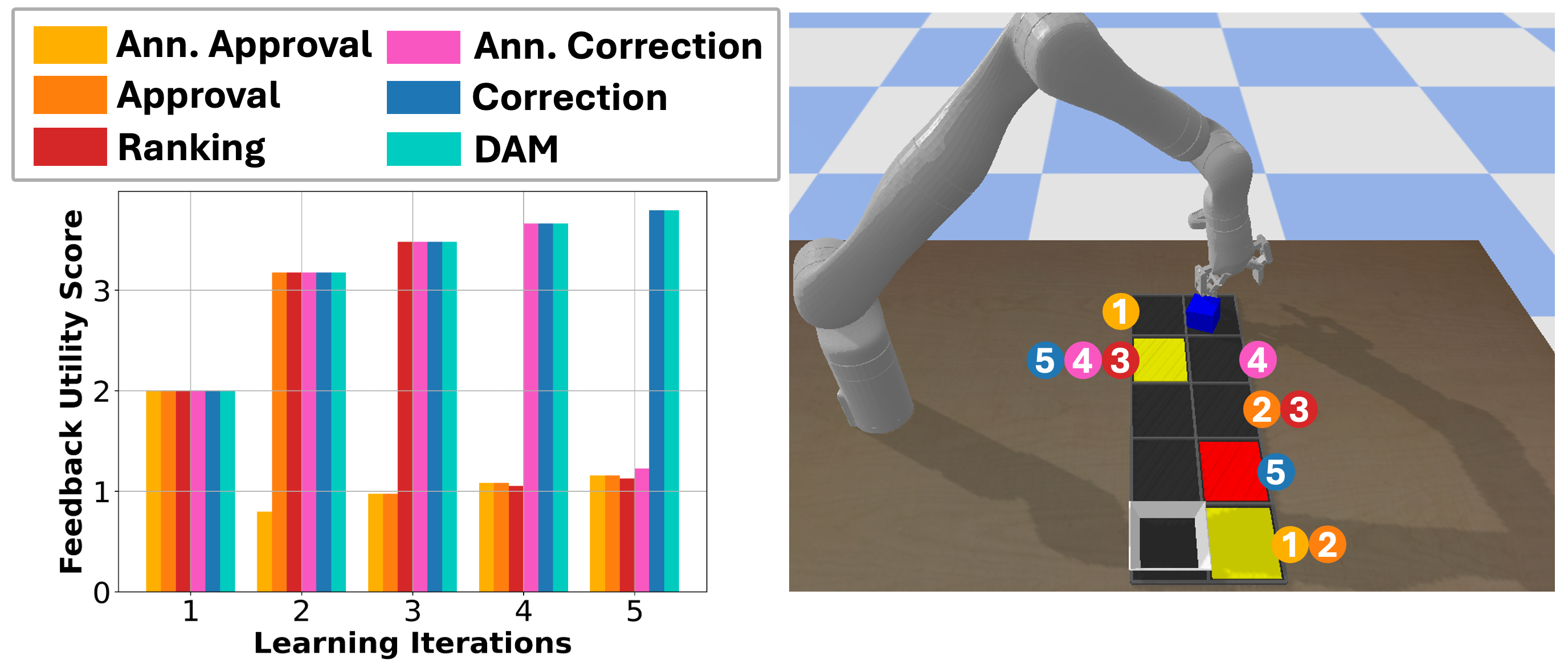}
        \caption{Feedback utility of each format across iterations. 
        Numbers mark when a state was identified as critical, and circle colors denote the chosen feedback format.}
        \label{fig:fb_val_progress}
\end{figure}

Figure~\ref{fig:fb_val_progress} illustrates the critical states and the most informative feedback formats selected at each iteration in the object delivery task using AFS, demonstrating that feedback utility changes over time, based on the robot's current knowledge.

\subsection{Stopping Criteria}
Besides guiding the selection of critical states and feedback format, the KL-divergence also serves as an indicator of when to stop querying. The querying phase can be terminated when $D_{KL}(\hat{p}^t \Vert \hat{q}^t)\leq\delta$, where $\delta$ is a problem-specific threshold. When  $D_{KL}(\hat{p}^t \Vert \hat{q}^t)\leq\delta$, it indicates that the learned model is a reasonable approximation of the underlying NSE distribution, and therefore the querying can be terminated even if the allotted budget $B$ has not been exhausted. The choice of $\delta$ provides a trade-off between thorough learning and human effort, and can be tuned based on domain-specific safety requirements.

\section{EXPERIMENTS IN SIMULATION}
\label{sec:ExpInSim}
We first evaluate AFS on three simulated domains (Fig.~\ref{fig:SimDomains}). Human feedback is simulated by modeling an oracle that selects safer actions with higher probability using a softmax action selection~\cite{ghosal2023effect,NEURIPS2020_2f10c157}: the probability of choosing an action $a'$ from a set of all safe actions $A^*$ in state $s$ is,
$\Pr(a' | s) = \frac{e^{Q(s, a')}}{\sum\limits_{a \in A^*} e^{Q(s, a)}}$.

\noindent \textbf{Baselines} (i) \emph{Naive Agent}: The agent naively executes its primary policy without learning about NSEs, providing an upper bound on the NSE penalty incurred.
(ii) \emph{Oracle}: The agent has complete knowledge about $R_T$ and $R_N$, providing a lower bound on the NSE penalty incurred.
(iii) \emph{Reward Inference with $\beta$ Modeling (RI)}~\cite{ghosal2023effect}: The agent selects a feedback format that maximizes information gain according to the human's inferred rationality, $\beta$.
(iv) \emph{Cost-Sensitive Approach}: The agent selects a feedback method with the least cost, according to the preference model $D$.
(v) \emph{Most-Probable Feedback}: The agent selects a feedback format that the human is most likely to provide, based on $D$.
(vi) \emph{Random Critical States}: The agent uses our AFS framework to learn about NSEs, except the critical states are sampled randomly from the entire state space.
We use $\theta_1\!=\!1$ and $\theta_2\!=\!1$ for all our experiments. AFS uses learned $\hat{R}_N$.

\noindent \textbf{Domains, Metrics and Feedback Formats} We evaluate the performance of various techniques on three domains in simulation (Figure ~\ref{fig:SimDomains}): outdoor navigation, vase and safety-gym's push. 
We optimize costs (negations of rewards) and compare techniques using average NSE penalty and average cost to goal, averaged over 100 trials.
For navigation, vase and push, we simulate human feedback.
The cost for $l_a$, $l_m$, and $l_h$ are $0$, $+5$, and $+10$ respectively.

\begin{figure}[t]
    \centering
    \includegraphics[width=0.8\linewidth]{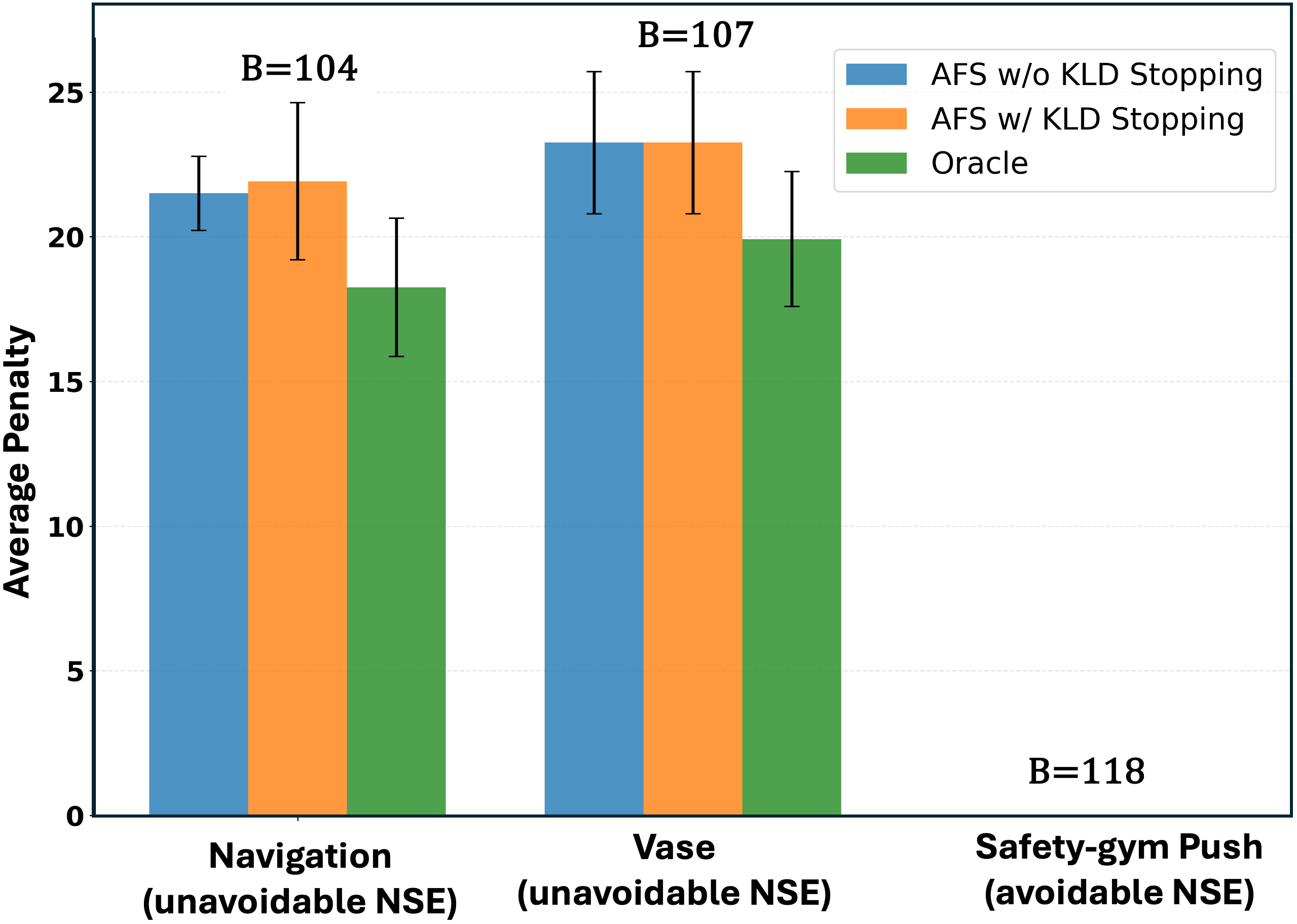}
    \caption{Average penalty incurred when learning with AFS using querying budget $B=400$, and KL divergence (KLD) as the stopping criterion. The budget utilized by AFS with KLD stopping is annotated in the plot.}
    \label{fig:stoppingCriteriaPenalty}
\end{figure}

\begin{figure*}[t]
    \centering
    \includegraphics[width=\textwidth]{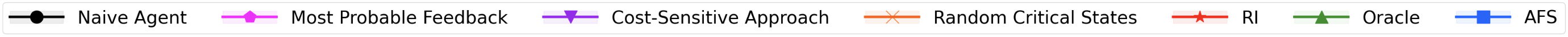} \\
    \includegraphics[width=\textwidth]{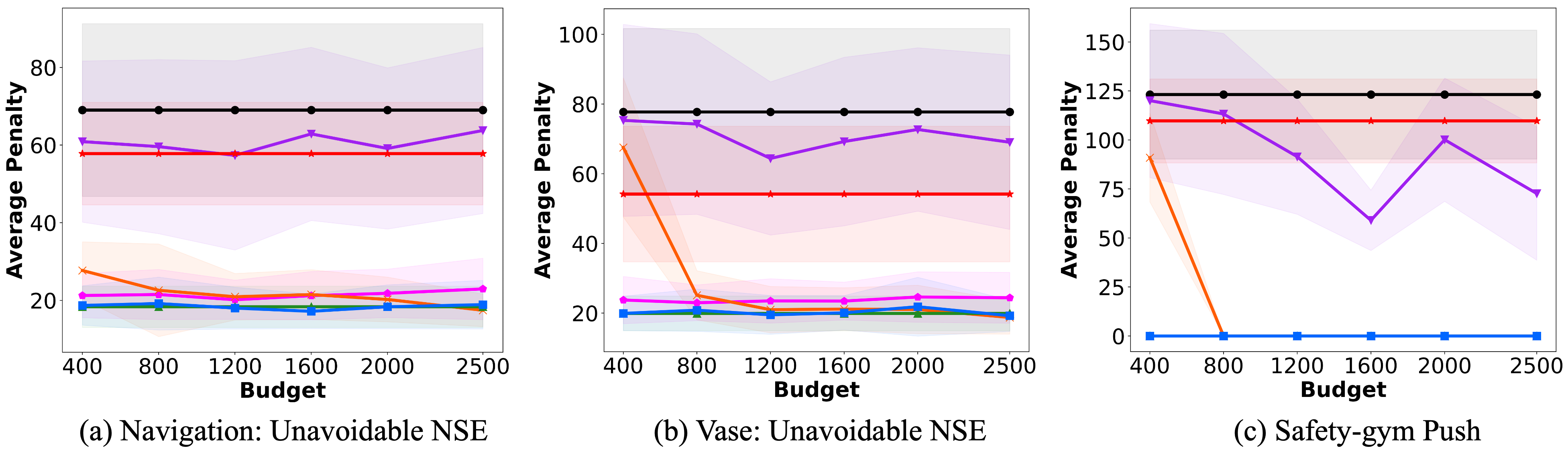}
    \caption{Average penalty incurred when querying with different feedback selection techniques.}
\label{fig:SimPenalties}
\end{figure*}

\noindent \textbf{Navigation:} In this ROS-based city environment, the robot optimizes the shortest path to the goal location. A state is represented as $\langle x, y, f, p \rangle$, where,
$x$ and $y$ are robot coordinates,
$f$ is the surface type (concrete or grass), and
$p$ indicates the presence of a puddle.
The robot can move in all four directions and each costs $+1$.  Actions succeed with probability $0.8$.
Navigating over grass damages the grass and is a mild NSE. Navigating over grass with puddles is a severe NSE. Features used for training are $\langle f, p \rangle$. Here, NSEs are unavoidable.

\noindent \textbf{Vase:} In this domain, the robot must quickly reach the goal, while minimizing breaking a vase as a side effect~\cite{krakovna2020avoiding}. A state is represented as $\langle x, y, v, c \rangle$ where,
$x$ and $y$ are robot's coordinates.
$v$ indicates the presence of a vase and
$c$ indicates if the floor is carpeted.
The robot moves in all four directions and each costs $+1$. Actions succeed with probability $0.8$. Breaking a vase placed on a carpet is a mild NSE
and breaking a vase on the hard surface is a severe NSE. $\langle v, c \rangle$ are used for training. All instances have unavoidable NSEs.

\noindent \textbf{Push:}
In this \texttt{safety-gymnasium} domain, the robot aims to push a box quickly to a goal state~\cite{ji2023safety}. Pushing a box on a hazard zone (blue circles) produces NSEs. We modify the domain such that in addition to the existing actions, the agent can also \emph{wrap} the box that costs $+1$. Every move action succeeds with probability 0.8, and the wrap action succeeds with probability 1.0. The NSEs can be avoided by pushing a wrapped box. A state is represented as $\langle x, y, b, w, h \rangle$ where, $x,y$ are the robot's coordinates, $b$ indicates carrying a box,  $w$ indicates if box is wrapped and $h$ denotes if it is a hazard area.
$\langle b, w, h \rangle$ are used for training.

\subsection{Results and Discussion}

\noindent\textbf{Effect of learning using AFS:} We first examine
the benefit of querying using AFS, by comparing the resulting
average NSE penalties and the cost for task completion, across domains and query budget.  Figure~\ref{fig:SimPenalties} shows the average NSE penalties when operating based on an NSE model learned using different querying approaches.
Clusters for critical state selection were generated using KMeans clustering algorithm with $K\!=\!3$ for navigation, vase and safety-gym's push domains (Figure~\ref{fig:SimPenalties} (a-c)).
The results show that our approach consistently performs similar to or better than the baselines.

There is a trade-off between optimizing task completion and mitigating NSEs, especially when NSEs are unavoidable. While some techniques are better at mitigating NSEs, they significantly impact task performance. Table~\ref{tab:baseline:reward} shows the average cost for task completion at $B\!=\!400$. \emph{Lower} values are better for both NSEs and task completion cost. While the Naive Agent has a lower cost for task completion, it incurs the highest NSE penalty as it has no knowledge of $R_N$.
RI causes more NSEs, especially when they are unavoidable, as its reward function does not fully model the penalties for mild and severe NSEs. Overall, the results show that our approach consistently mitigates avoidable and unavoidable NSEs, without affecting the task performance substantially.

Figure~\ref{fig:stoppingCriteriaPenalty} shows the average penalty when AFS uses KL-divergence (KLD) as the stopping criteria, compared to querying with budget $B=400$. For comparison, we also annotate in the plot the querying budget used by AFS with KLD stopping at the time of termination. The results show that despite terminating earlier and using few queries, AFS with the KLD stopping achieves comparable performance to that of AFS with query budget $B=400$, demonstrating the usefulness of KLD as a stopping criterion.

\begin{table}[ht]
    \vspace{5pt}
    \centering
    \small
    \setlength{\tabcolsep}{3pt}
    \begin{tabular}{|p{1.5cm}|p{2cm}|p{2cm}|p{2cm}|} \hline
         Method &  Navigation: \mbox{unavoidable} NSE&  Vase: \mbox{unavoidable} NSE& Safety-gym Push: \mbox{avoidable} NSE \\
         \hline \hline
         Oracle& $51.37\pm2.69$ & $54.46\pm6.70$ & $44.62\pm9.97$\\
         Naive&  $36.11\pm1.39$&  $36.0\pm2.89$ & $39.82\pm5.44$\\
         RI &  $40.10\pm0.69$ &  $37.42\pm1.01$ & $42.15\pm2.44$\\
         AFS &$64.8\pm2.3$ & $52.68\pm7.87$ & $48.32\pm4.42$\\
         \hline
    \end{tabular}
    \caption{Avg. cost and standard error at task completion.}
    \label{tab:baseline:reward}
\end{table}

\section{In-Person User Study with a Physical Robot Arm}
We conducted an in-person study with a Kinova Gen3 7DoF arm~\cite{kinova} tasked with delivering two objects---an orange toy and a white box---across a workspace containing items of varying fragility (Figure~\ref{fig:userStudySetup}). This setup involves 
users providing both interface-based and kinesthestic feedback to the robot. The study was approved by Oregon State University IRB. Participants were compensated with a $\$15$ Amazon gift card for their participation in the study.

This user study had three goals: (1) to measure our approach's effectiveness in reducing NSEs for a real-world task, (2) to understand how users perceive the adaptivity, workload and competence of the robot operating in the AFS framework, and (3) to evaluate the extent to which AFS captures user preferences in practice, while ensuring maximum information gain during the learning process.

\subsection{Methods}

\subsubsection{Participants}
We conducted a pilot study in simulation to inform our overall design, the details of which are discussed in Appendix~\ref{sec:StudyInSim}. We  conducted another pilot study with $N\!=\!10$ participants to evaluate the study setup with the Kinova arm. In particular, this pilot study assessed the clarity of instructions, survey wording, and feasibility of the task design in the object delivery task of the Kinova arm. Based on the participant feedback, we simplified the survey questions and included example trajectories that demonstrated safe and NSE-causing behaviors.
For the main study, we recruited $N\!=\!30$ participants with basic computer literacy from the \emph{general population} through university mailing lists and public forums. Participants were aged $18$--$72$ years ($M\!=\!32.10, SD\!=\!13.11$), with $53.3\%$ men and $46.7\%$ women. Participants reported varied prior experience with robots: $73.3\%$ had general awareness of similar robot products, $6.7\%$ had researched or investigated robots, $3.3\%$ had interacted through product demos, and $13.3\%$ had no prior awareness of similar products.

\subsubsection{Robotic System Setup} 
\begin{figure*}[t]
    \centering
    \includegraphics[width=\textwidth]{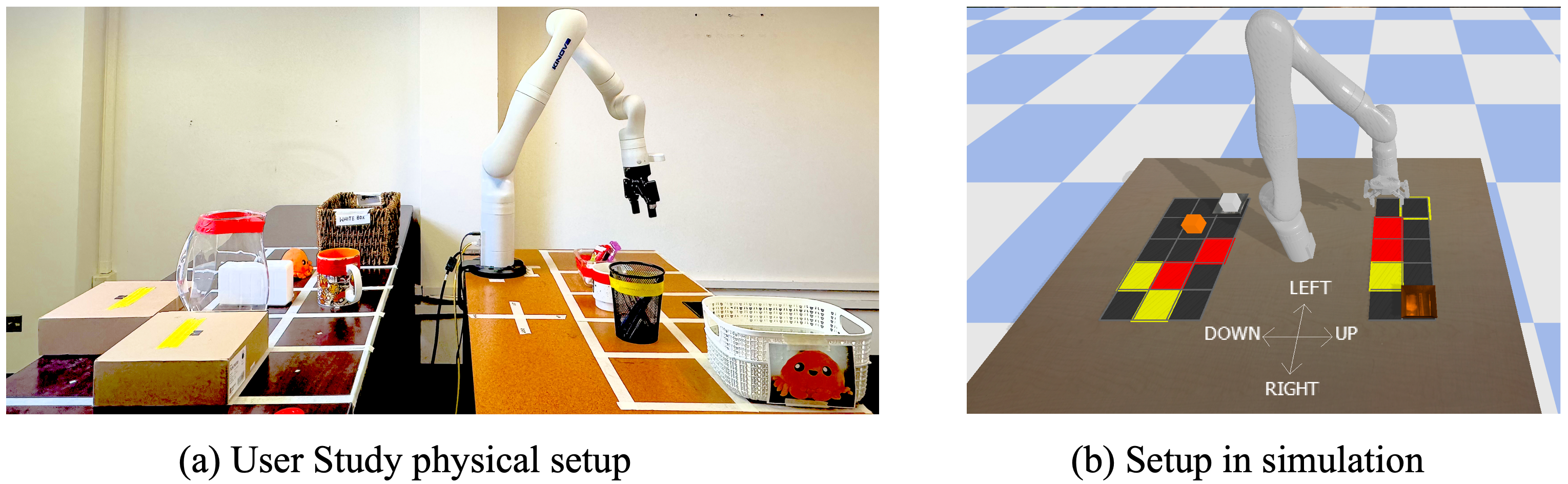}
    \caption{Task setup for the human subject study. \textbf{(a)} Physical setup of the task for human subjects study; \textbf{(b)} Replication of the physical setup using PyBullet. A dialog box corresponding to the current feedback format is shown for every query.}
  \label{fig:userStudySetup}
\end{figure*}

The Kinova Gen3 arm was equipped with a joint space compliant controller which allowed participants to physically move the joints of the arm through space with
gravity compensation when needed. Additionally, a task-space planner allowed for navigation to discrete grid positions for both feedback queries and policy execution~\cite{kinova}.  
Figure~\ref{fig:userStudySetup}(a) shows the physical workspace and the two delivery objects, while Figure~\ref{fig:userStudySetup}(b) shows the corresponding PyBullet simulation used for visualization during GUI-based feedback. A dialog box was displayed to prompt the participant whenever feedback was queried\footnote{See Appendix~\ref{sec:3} for details on the dialog box and examples for each feedback format.}.

\subsubsection{Interaction Premise}
The interaction simulated an assistive robot delivering objects to their designated bins. Specifically, the task required the Kinova arm to deliver an orange plush toy and a rigid white box to their respective bins while avoiding collision with surrounding obstacles of different fragility. Collisions with fragile obstacles (e.g. a glass vase) during delivery of the plush toy were considered a mild NSE. Collisions involving the white rigid box were severe NSEs if with a fragile object and were mild NSEs if with a non-fragile object. All other scenarios were considered safe. The workspace was discretized into a grid of cells marked with tape on the tabletop and mirrored in the GUI. Each cell represented a state corresponding to possible end-effector position.

\subsubsection{Study Design}
The robot's state space was discretized and represented as $\langle x, y, i_1, i_2, o, f, g_1, g_2\rangle$, where $(x, y)$ denote the end-effector position, $i_1$ and $i_2$ indicate the presence of either orange plush toy or white rigid box in the end effector, $o$ indicates the presence of an obstacle, and $f$ indicates obstacle fragility, and $g_1$ and $g_2$ indicate whether either of the objects were delivered in their corresponding goal locations (i.e., orange plush toy in white bin and the white box in the wicker bin). 

Participants interacted with the robot through \emph{four} feedback formats, $\mathcal{F} = \{\text{App, Corr, Rank, DAM}\}$, during both the training and main experience phases. 
Depending on the feedback format, the Kinova arm executed the queried action in the physical workspace or displayed a simulation of the action in the graphical user interface (GUI). Interaction across the four feedback formats are described below.

\begin{enumerate}
    \item \textbf{Approval:} The robot executed a single action in simulation, and participants indicated whether it was safe by selecting ``yes" or ``no" in the GUI.
    \item \textbf{Correction:} The robot first executes action prescribed by its policy in simulation. If the action in simulation is deemed unsafe by the participant, the robot in the physical setup moves to the queried location. Participants then correct the robot by physically moving the robot arm to demonstrate a safe alternative action.
    \item \textbf{Demo-Action Mismatch:} The robot first physically moved its arm to a specific end-effector position in the workspace. Participants then provided feedback by guiding the arm to a safe position, thereby demonstrating the safe action. The robot compares the action given by its policy to the demonstrated action. If the robot's action and the demonstrated actions do not match, then the robot's action is considered unsafe.
    \item \textbf{Ranking:} Simulation clips of two actions selected at random in a given state were presented in GUI. Participants compared the two candidate actions and selected which was safer. If both actions were judged equally safe or unsafe, either option could be chosen.
\end{enumerate}

Each participant experienced four learning conditions in a within-subjects, counterbalanced design:
\begin{enumerate}
    \item The baseline RI approach proposed in~\cite{ghosal2023effect}, 
    \item AFS with random $\Omega$, where critical states are randomly selected,
    \item AFS with a fixed feedback format (DAM) for querying, consistent with prior works that rely primarily on demonstrations, and
    \item The proposed AFS approach, where both the feedback format and the critical states are selected to maximize information gain.
\end{enumerate}

Each condition is a distinct feedback query selection strategy controlling how the robot queried participants during learning. These conditions are the independent variables. The dependent measures include NSE occurrences, their severity, perceived workload, trust, competence and user alignment.

\subsubsection{Hypotheses}
We test the following hypotheses in the in-person study. These hypotheses were derived from trends observed in the experiments and human subjects study in simulation (Sec. ~\ref{sec:ExpInSim} and Appendix~\ref{sec:StudyInSim}). 

\textbf{H1:} \emph{Robots learning using AFS will have fewer NSEs in comparison to the baselines.} \\ This hypothesis is derived from the results of our experiments on simulated domains (Figure~\ref{fig:SimPenalties}) where AFS consistently reduced NSEs while completing the assigned task. We hypothesize that this trend extends to physical human-robot interactions. 

\textbf{H2:} \emph{AFS will achieve comparable or better performance compared to the baselines, with a lower perceived workload for the users.} \\ The results on simulated domains (Figure~\ref{fig:stoppingCriteriaPenalty}) show that AFS achieved better or comparable performance to the baselines, using fewer feedback queries. While the in-person user study requires relatively greater physical and cognitive effort, we expect the advantage of the sample efficiency to persist and  investigate whether it translates to reduced perceived workload.

\textbf{H3:} \emph{Participants will report AFS as more trustworthy, competent, and aligned with user expectations, in comparison to the baselines.} \\ In the human subjects simulation study (Table \ref{tab:user_study_qual} in the Appendix), participants reported that AFS selected intelligent queries, targeted critical states, and improved the agent's performance, reflecting indicators of trust, competence and user alignment. We hypothesize that this trend extends to physical settings as well.

Hypotheses \textbf{H1} and \textbf{H2} explore trends identified in simulation and are therefore confirmatory. Hypothesis \textbf{H3} builds on the perception measures used in the human subjects study in simulation, and is hence treated as an extended confirmatory hypothesis.

\subsubsection{Procedure}
Each study session lasted approximately one hour and followed three phases. 
\paragraph{Training} Participants were first introduced to the task objective, workspace, and the four feedback formats. For each format, they provided feedback on four sample queries to practice both GUI-based and kinesthetic interactions. After the completing each format, the participants rated the following: (i) probability of responding to a query in that format, $\psi(f)$, (ii) perceived cost or effort required to provide feedback, $C(f)$, and (iii) the overall task workload. This phase helped establish measures like feedback likelihood, perceived effort, and workload.
\paragraph{Main Experience} Following training, participants completed the four learning conditions corresponding to different approaches under evaluation. In each condition, the participants provided feedback to train the robot to avoid collision while performing the object-delivery task. Depending on the feedback format selected by the querying strategy, participants either evaluated short simulation clips on the GUI or physically guided the robotic arm. At the end of each condition, the robot executed its learned policy based on its learning under that condition. The participants then observed its performance and completed a brief post-condition questionnaire assessing workload, trust, perceived competence, and user-alignment.

\paragraph{Closing} At the end of the study, participants compared the four learning approaches in terms of trade-offs between learning speed and safety. Participants reported their preferences on providing feedback through multiple formats versus relying on a single feedback format. These responses offered qualitative insight into AFS's practicality and user acceptance.

\begin{figure*}[ht]
    \centering
    \includegraphics[width=0.9\textwidth]{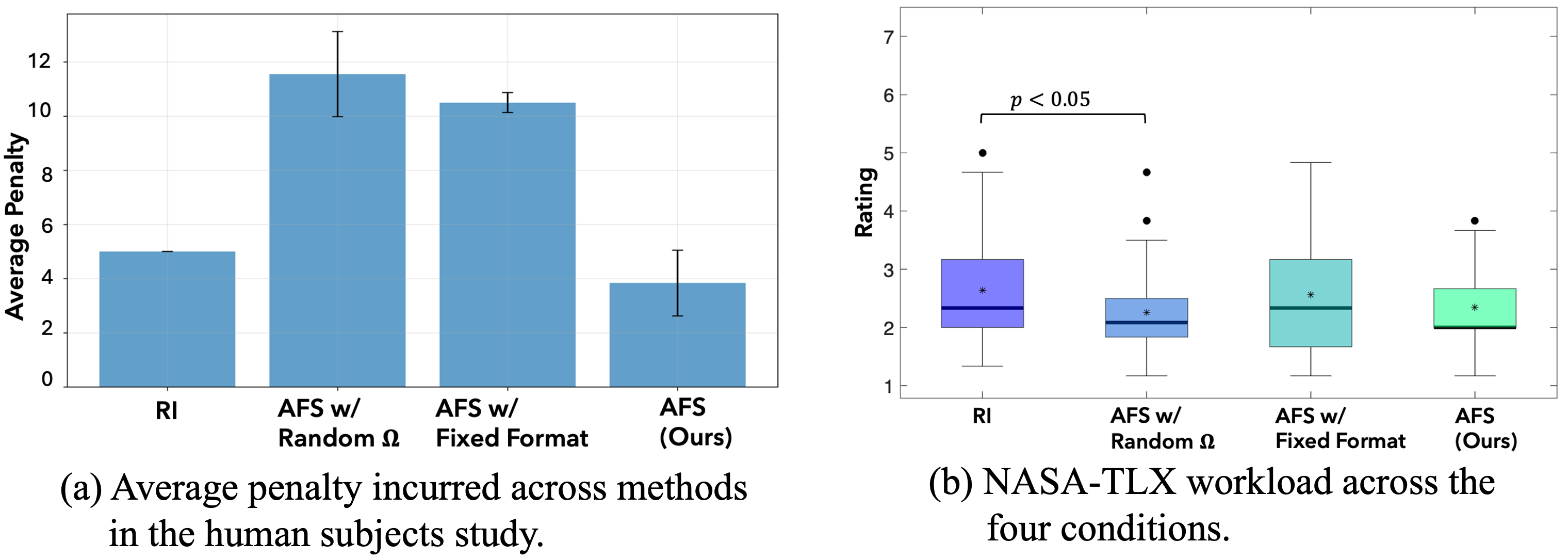}
    \caption{Results from the user study on the Kinova 7DoF arm.}
\label{fig:UserStudyKinova}
\end{figure*}

\subsubsection{Measures}
We collected both quantitative and qualitative measures. The quantitative measure captured task-level performance through the frequency and the severity of NSEs (mild and severe). Qualitative measures captured participants' perceptions of the following. 
\begin{enumerate}
    \item \textbf{Workload:} Participants' perceived workload across the feedback formats and learning conditions were measured using the NASA Task Load Index (NASA TLX)~\cite{Hart1988DevelopmentON}.
    The questionnaire scales were transformed to 7-point subscales ranging from ``Very Low" (1) to ``Very High" (7). Responses were collected during the training phase and after each condition in the main experience phase.
    \item \textbf{Robot Attributes:} Perceived robot attributes, like competence, warmth and discomfort, were measured using the 9-point Robotic Social Attributes Scale (RoSAS)~\cite{8534914}, ranging from ``Strongly Disagree" (1) to ``Strongly Agree" (9). Participants completed this questionnaire after each learning condition. 
    \item \textbf{Trust:} A custom 10-point trust scale $(0\%-100\%)$ was used to measure participants' confidence in the robot's ability to act safely under each learning condition. Participants rated their trust both before and after the robot's training phase to capture changes in its learning performance. 
    \item \textbf{User Alignment:} Participants' perception of user alignment was assessed using a custom 7-point Likert scale ranging from ``Strongly Disagree" (1) to ``Strong Agree" (7). Participants rated (i) how well the critical states queried by the robot aligned with their own assessment of which states were important for learning, and (ii) how well the feedback formats chosen across conditions matched their personal feedback preferences. Higher rating indicated stronger perceived alignment between the robot's querying strategy and the participants' expectations.
\end{enumerate}

\subsubsection{Analysis}
Survey responses were compiled into cumulative RoSAS (competence, warmth, discomfort) and NASA-TLX workload scores. A repeated-measures ANOVA (rANOVA) tested for significant differences across learning conditions; we report the $F$-statistic, $p$-value and effect size as generalized eta-squared ($\eta^2_G$). When effects were significant, Tukey's post-hoc tests identified pairwise differences. 
All results are reported with means (M), standard errors (SE), and $p$-values.

\subsection{Results}
We evaluate hypotheses \textbf{H1-H3} using both objective and subjective measures. Data from all 30 participants were included in the analysis, as all sessions were completed successfully. 

\subsubsection{Effectiveness of AFS in Mitigating NSEs (Hypothesis \textbf{H1})}
Figure~\ref{fig:UserStudyKinova}(a) shows the average penalty incurred under each condition. AFS approach incurred the least NSE penalty ($M=3.83, SE=1.21$), 
substantially lower than AFS with random $\Omega$ ($M=11.55, SE=1.57$) and AFS with a fixed feedback format ($M=10.50, SE=0.37$). The RI baseline incurred higher penalties ($M=5.00, SE=0.00$) compared to AFS. These results confirm hypothesis \textbf{H1} and demonstrate that adaptively selecting both critical states and feedback formats reduced unsafe behaviors more effectively than random or fixed querying strategies.

\begin{figure*}[t]
    \centering
    \includegraphics[width=1\textwidth]{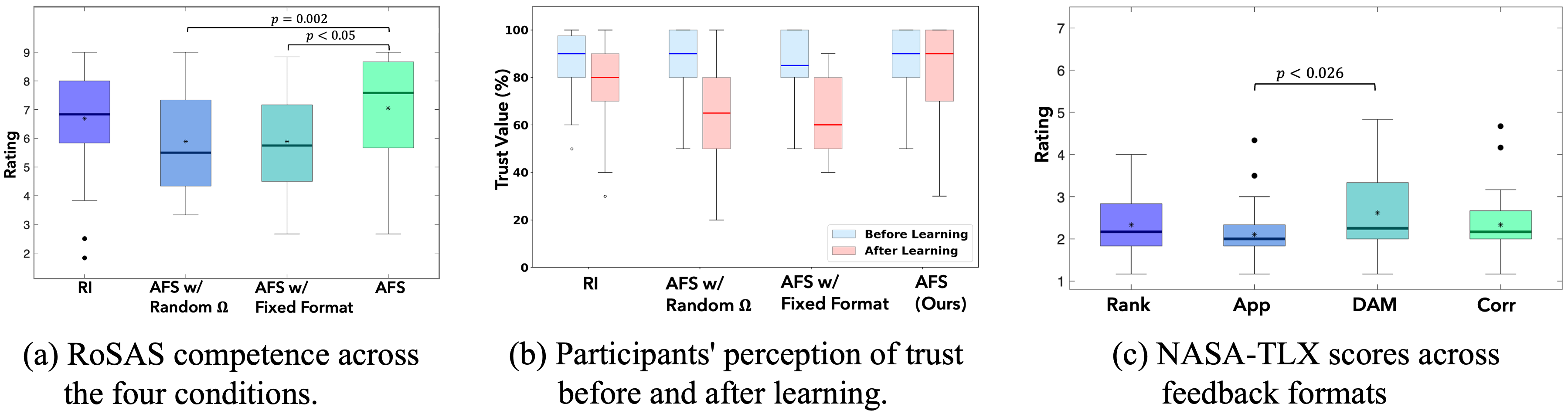}
    \caption{User study results. \textbf{(a)-(b)} RoSAS competence and NASA Task-Load across the four conditions in the main study; \textbf{(c)} NASA Task-Load across feedback formats.}
  \label{fig:userStudyResults}
\end{figure*}

\subsubsection{Learning Efficiency and Workload (Hypothesis \textbf{H2})}
We first compare the perceived workload across different feedback formats, followed by the results across learning conditions. 
Demonstration is the most widely used feedback format in existing works but was perceived as the most demanding (Figure~\ref{fig:userStudyResults}(c)). 
While corrections offer corrective action in addition to disapproving agent's action, it also imposed substantial effort on the users.
Approval required the least workload but conveyed limited information. A repeated-measures ANOVA revealed a significant effect of feedback format on perceived workload, ($F(3,87)=3.33, p=0.023, \eta_G^2=0.046$). Post hoc comparisons indicated that Approval ($M=2.11, SE=0.12$) imposed significantly lower workload ($p=0.026$) than Demo-Action Mismatch ($M=2.62, SE=0.19$), while no other pairwise differences reached significance. This trade-off underscores the need for an adaptive selection strategy to balance informativeness with user effort. 

The rANOVA analysis across the four learning conditions further revealed a significant effect in the NASA-TLX workload ratings ($F(3, 87)=3.73, p=0.014, \eta_G^2=0.030$). Among the four conditions, AFS achieved one of the lowest perceived workload ratings ($M = 2.34, SE = 0.12$), comparable to AFS with random $\Omega$ ($M = 2.26, SE = 0.15$) and lower than both AFS with fixed format ($M = 2.56, SE = 0.19$) and RI ($M = 2.64, SE = 0.19$). Tukey post-hoc tests showed that workload in AFS with random $\Omega$ imposed a significantly lower workload than RI ($p=0.033$). Overall, these results support \textbf{H2}, indicating that adaptively selecting queries helps reduce perceived workload relative to the baselines (Figure~\ref{fig:UserStudyKinova}(b)).

\subsubsection{Trust, Competence, and Preference Alignment (Hypothesis \textbf{H3})}

Participants' rating on the robot's ability to act safely increased after learning with AFS, as shown in Figure~\ref{fig:userStudyResults}(b). A significant effect was also found for perceived robot competence ($F(3,87)=10.6, p<0.001, \eta_G^2=0.082$) (Figure~\ref{fig:userStudyResults}(a)). AFS was rated highest ($M = 7.04, SE = 0.32$), significantly greater than AFS with random $\Omega$ ($M = 5.88, SE = 0.32, p=0.002$) and AFS with fixed format ($M = 5.88, SE = 0.30, p<0.001$), while comparable to RI ($M = 6.68, SE = 0.32$). These results support \textbf{H3}---AFS was perceived as more competent and trustworthy compared to the baselines.

Descriptive analyses of user alignment on state criticality and feedback alignment ratings, indicated consistent trends across participants. While differences between conditions were not statistically significant ($p>0.05$), AFS consistently received higher ratings for feedback alignment ($M = 3.79, SE = 0.42$) relative to state criticality ($M = 3.14, SE = 0.40$), suggesting that participants found AFS's query selections relevant and aligned with their preferences. Participants (both those aware and unaware of similar robotic systems) perceived AFS's queries as critical for learning and well-aligned with their feedback preferences. Participants with prior research experience rated state criticality and format alignment comparable, indicating confidence in adaptivity of AFS's querying process.

\section{Discussion}
Our experiments followed an increasingly realistic progression in design. In the experiments in simulation with both avoidable and unavoidable NSEs, AFS incurred lower penalties and overall costs compared to the baselines, demonstrating its ability to balance task performance with safety. The results of our pilot study, where users interacted with a simulated agent, showed that AFS effectively learns the participant's feedback preference model and uses them to select formats aligned with user expectations. Finally, the in-person user study with the Kinova arm, showed the practicality of using AFS in real-world settings, achieving favorable ratings on trust, workload, and user-preference alignment.
These finding support our three hypotheses regarding the performance of AFS: (H1) it reduces unsafe behaviors more effectively than the baselines, (H2) it improves learning efficiency while reducing user workload, and (H3) it is perceived as more trustworthy and competent. The results collectively highlight that adaptively selecting both the query format and the states to pose the queries to the user enhances learning efficiency and reduces user effort.

Beyond confirming these hypotheses, the findings provide important design implications for human-in-the-loop learning systems. By modeling the trade-off between informativeness and effort, AFS offers a framework to balance user workload with the need for high-quality feedback. The learned feedback preference model allows the agent to adaptively select querying formats while minimizing human effort.
Using KL-divergence as stopping criterion further enables adaptive termination of the querying process.
This overcomes the problem of determining the ``right'' querying budget for a problem, and shows that AFS enables efficient learning while minimizing redundant human feedback. 
These design principles can inform the development of interactive systems that adapt query format and frequency based on agent's current knowledge and user feedback preferences.
Overall the results show that AFS (1) consistently outperforms the baselines across different evaluation settings, and (2) can be effectively deployed in real-world human-robot interaction scenarios. 

A key strength of this work lies in its extensive evaluation, from simulation to real robot studies, supporting AFS's robustness and practicality. One limitation, however, is that the current evaluation focuses on discrete environments. Extending AFS to continuous domains introduces challenges such as identifying critical states and estimating divergence-based information gain in high-dimensional spaces. While gathering feedback at the trajectory-level is relatively easier in continuous settings, gathering state-level feedback, which is the focus of this work, is challenging. These challenges stem from the need for scalable state representations and efficient sampling strategies, which will be a focus for future work.

\section{Conclusion and Future Work}
The proposed Adaptive Feedback Selection (AFS) facilitates querying a human in different formats in different regions of the state space, to effectively learn a reward function. Our approach uses information gain to identify critical states for querying, and the most informative feedback format to query in these states, while accounting for the cost and uncertainty of receiving feedback in each format.
Our empirical evaluations using four domains in simulation and a human subjects study in simulation demonstrate the effectiveness and sample efficiency of our approach in mitigating avoidable and unavoidable negative side effects (NSEs). 
The subsequent in-person user study with a Kinova Gen3 7DoF arm further validates these finding, showing that AFS not only improves NSE avoidance but also enhances user trust, competence perception, and user-alignment.
While AFS assumes that human feedback reflects a true underlying notion of safety, biased feedback can misguide the robot and lead to unintended NSEs. Understanding when such biases arise and how to correct for them remains an open challenge. Extending AFS with bias-aware inference mechanisms is a promising future direction. Future work will also focus on extending AFS to continuous state and action spaces, strengthening AFS's applicability to complex, safety-critical domains where user-aware interaction is essential.


\section*{ACKNOWLEDGMENT}

This work was supported in part by National Science Foundation grant number 2416459.


\bibliographystyle{IEEEtran}
\bibliography{references}

\newpage

\appendices
This appendix presents additional simulation results, discusses the human subjects pilot study in simulation, and provides additional details on the in-person user study with a Kinova arm.

\section{Additional Experiments in simulation}
\label{sec:1}

\begin{figure*}[ht]
    \centering
    \includegraphics[width=\textwidth]{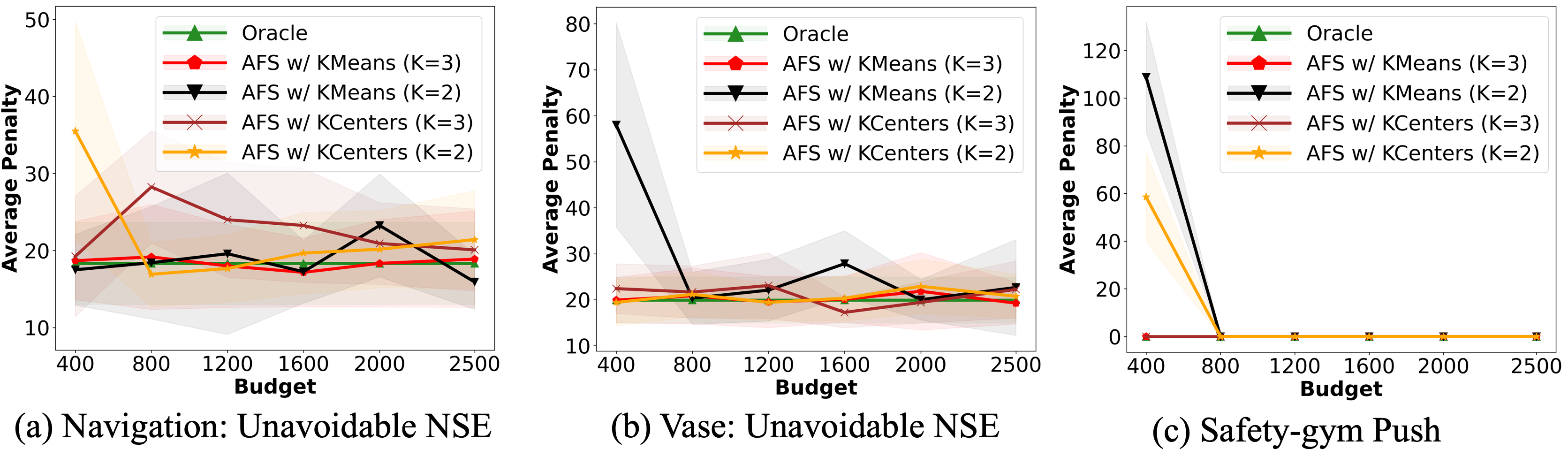}
    \caption{Average penalty incurred using our approach (AFS) with KMeans and KCenters clustering algorithm, evaluated across varying number of clusters ($K$).}
\label{fig:Diff_clusters}
\end{figure*}

\begin{figure*}[ht]
    \centering
    \includegraphics[width=\linewidth]{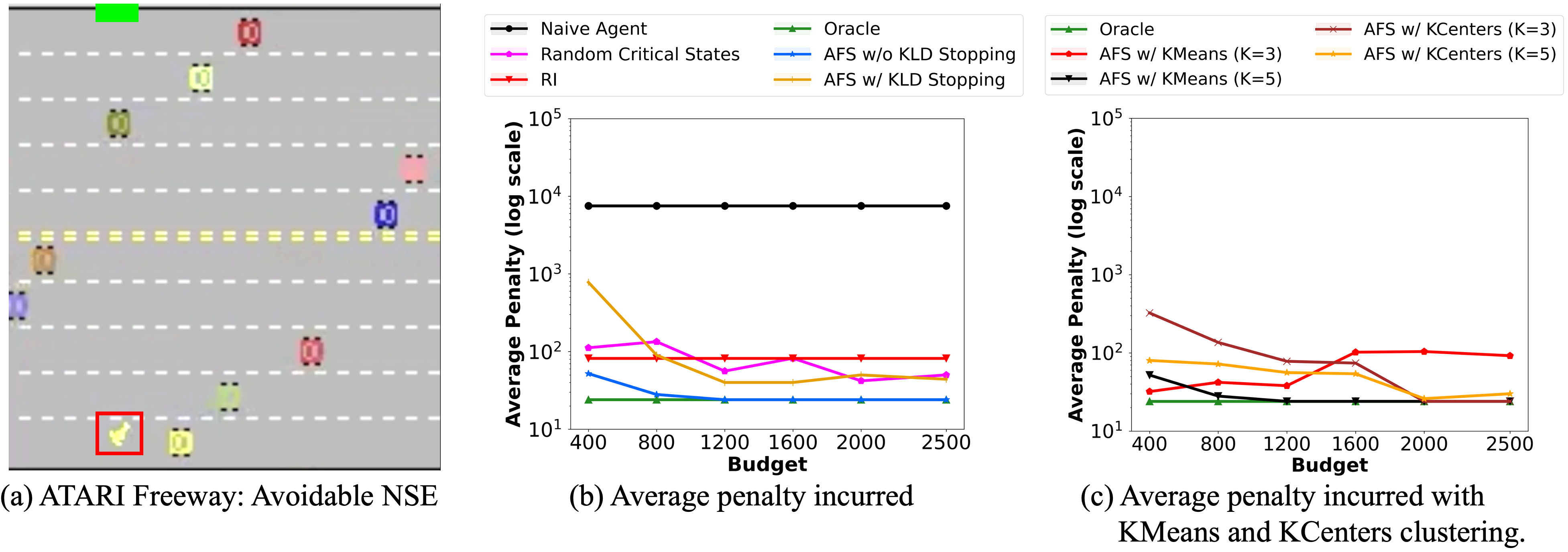}
    \caption{An instance of the Freeway domain, and the average penalty incurred.}
    \label{fig:atari_penalty}
\end{figure*}

\subsection{Effect of Clustering} 
To understand how clustering influences the effectiveness of our approach (AFS), we evaluate AFS under different clustering configurations. In particular, we vary the number of clusters and the clustering approach. 

Figure~\ref{fig:Diff_clusters} shows the average penalty incurred using AFS with the KMeans and KCenters clustering algorithms for varying numbers of clusters ($K\!=\{2,3\}$ in the navigation, vase and push domains). We restrict our evaluation to these $K$ values since the maximum number of distinct clusters in each domain is determined by number of unique combinations of state features.
In the navigation domain, features used for clustering states are $\langle f, p\rangle$. The valid unique combinations are $\langle f=$ concrete, $p=$ no puddle$\rangle$, $\langle f=\text{grass}, p=\text{no puddle}\rangle$, and $\langle f=\text{grass}, p=\text{puddle}\rangle$. Hence, having $K>3$ will not produce unique clusters.
Similarly, in the vase domain, features used for clustering are $\langle v, c \rangle$, where the unique, valid combinations are $\langle \text{no vase, no carpet}\rangle, \langle\text{vase, no carpet}\rangle, \langle\text{vase, carpet}\rangle$. For the push domain, the features used for clustering are $\langle b, w, h\rangle$, with valid unique combinations including $\langle\text{no box}, \text{not wrapped}, \text{hazard}\rangle$, $\langle\text{box}, \text{not wrapped}, \text{hazard}\rangle$, $\langle\text{no box}, \text{not wrapped}, \text{no hazard}\rangle$, $\langle\text{box}, \text{wrapped}, \text{no hazard}\rangle$. 
The results in Figure~\ref{fig:Diff_clusters} demonstrate that increasing $K$ generally improves the performance of our approach, with both clustering methods. A higher number of clusters allows for a more refined grouping of states based on distinct state features, enabling the agent to query the human for feedback across a more diverse set of states. This diversity enhances the agent's ability to accurately learn and mitigate NSEs.

\subsection{Learning with Implicit and Explicit Feedback Formats}

To demonstrate AFS's flexibility across feedback modalities, we extend it to handle both explicit (e.g., DAM) and implicit (e.g., gaze) feedback formats.

\subsubsection{Implicit Feedback Format}
We consider a human's gaze on the screen as the implicit feedback format. Here, the robot requests to collect gaze data of the user and compares its action outcomes with the gaze positions of the user~\cite{saran2021efficiently}. Actions with outcomes aligning with the average gaze direction are labeled as acceptable ($l_a$), and unacceptable ($l_h$) otherwise.

\subsubsection{Domain}

We evaluate AFS in the Atari Freeway environment, where the robot (a chicken) navigates ten cars moving at varying speeds to reach the destination quickly while avoiding being hit(Figure~\ref{fig:atari_penalty}(a)).
Being hit by a car moves the robot back to its previous position, and is a severe NSE.
A game state is defined by coordinates $(x_1, y_1)$ and $(x_2, y_2)$, i.e., the top left and bottom right corners of the robot and cars, extracted from the Atari-HEAD dataset~\cite{zhang2020atari}. Similar to \cite{saran2021efficiently}, only car coordinates within a specific range of the robot are considered.
The robot can move up, down or stay in place, with unit cost and deterministic transitions.

\subsubsection{Effect of Learning using AFS}
Figure~\ref{fig:atari_penalty}(b) shows the average NSE penalties when operating based on an NSE model learned using different querying approaches. Clusters for critical state selection were generated using KMeans clustering algorithm with $K\!=\!5$ in the Atari Freeway domain. 

\begin{table}[ht]
    \centering
    \begin{tabular}{c|c}
        \toprule
        \textbf{Method} & \textbf{Avg. Cost} \\
        \midrule
         Oracle & $3759.8 \pm 0.00$ \\
         Naive & $61661.0 \pm 0.00$ \\
         RI &  $71716.6 \pm 0.00$ \\
         AFS (Ours) & $1726.5 \pm 0.00$\\
         \bottomrule
    \end{tabular}
    \caption{Average cost at task completion.}
    \label{tab:atari_cost}
\end{table}

Table~\ref{tab:atari_cost} shows the average cost for task completion. While the Naive Agent has a lower cost for task completion, it incurs the highest NSE penalty as it has no knowledge of $R_N$. RI causes more NSEs, as its reward function does not fully model the penalties for mild and severe NSEs. Overall, the results show that AFS consistently mitigates NSEs, without affecting the task performance substantially.

\section{Human Subjects Pilot Study in Simulation}
\label{sec:StudyInSim}

\begin{figure}[ht]
    \centerline{
    \includegraphics[width=\linewidth]{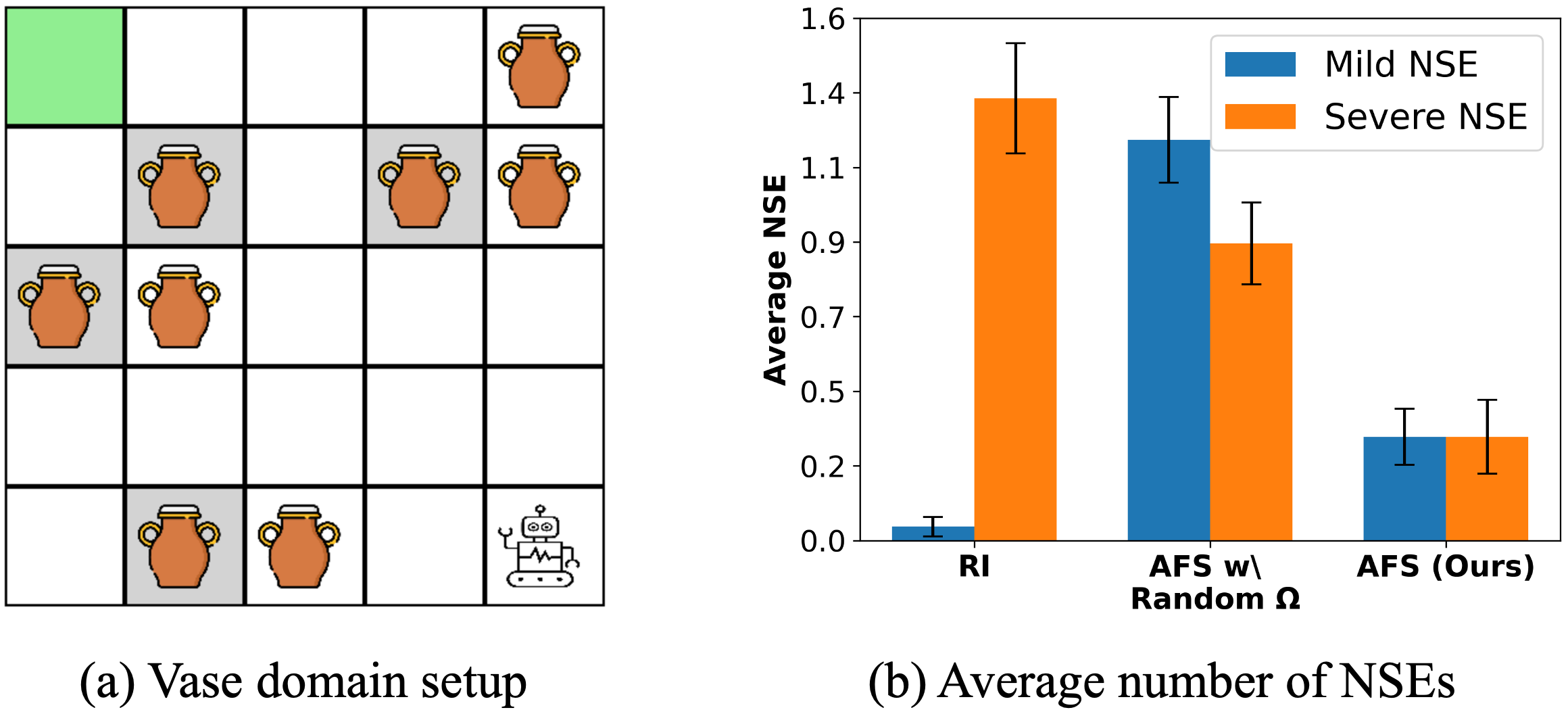}}
  \caption{Results from the user study on a simulated domain.}
\label{fig:user_study_domain}
\end{figure}

We conducted a within-subjects pilot study on a $5\!\times\!5$ Vase domain in simulation as shown in Fig.~\ref{fig:user_study_domain}(a), with $12$ human participants who had completed at least one course in Reinforcement Learning. The objective of this study is to evaluate whether: (1) AFS outperforms the baselines when a feedback preference model is learned from user interactions; (2) the selected feedback formats and critical states enhance agent's learning, and align with user preferences. The study was conducted with approval from Oregon State University IRB, and the participants were compensated with a $\$10$ Amazon gift card for their time.

\subsection{Study Design}
After introducing the domain and the agent's objective, users completed a tutorial where they interacted with the system by providing feedback in each of the six formats. The study interface included feedback buttons that varied based on the format. This was followed by a calibration phase, during which the users' preference model was learned. Each user was prompted five times per format to provide feedback, with the option to respond or ignore, allowing them to express their interaction preferences. The probability of receiving feedback in a given format was determined by the fraction of prompts the user responded to, while the cost was based on their self-reported effort.

The study comprised three phases, each evaluating a different baseline approaches to select feedback queries: (1) RI, (2) AFS with Random $\Omega$, and (3) AFS with our proposed method for critical state selection. To prevent bias, users were unaware of the approach used in each phase. After completing a phase, they were shown a trajectory of the agent’s learned policy and asked to evaluate the approach used in that phase.

\subsection{User Interface for Feedback Collection}

\begin{figure*}[t]
    \centering
    \includegraphics[scale=0.5]{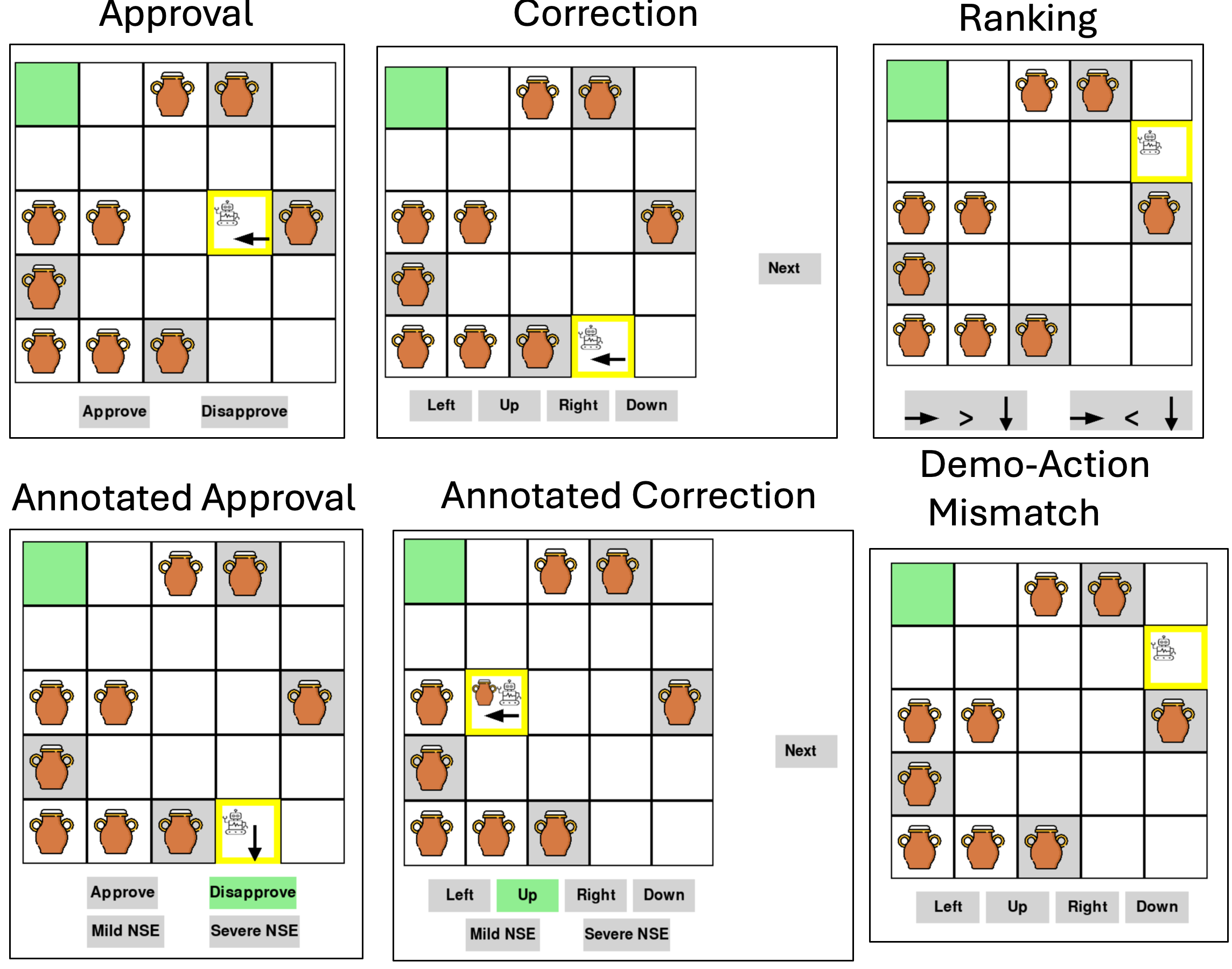}
    \caption{Interface for the human subjects study in simulation. Participants provide feedback via button clicks, with available options varying by format, as shown.}
    \label{fig:userStudyInSim_interface}
\end{figure*}

Figure~\ref{fig:userStudyInSim_interface} illustrates the interface used in the simulation-based human subjects study. Participants interacted with the simulated robot through a GUI consisting of feedback buttons whose labels and available options varied depending on the feedback format. For each query, the interface would display the action the agent intends to take in the gridworld and provide a corresponding set of input buttons to record user feedback. 

\begin{table*}[t]
    \vspace{5pt}
    \centering
    \small
    \setlength{\tabcolsep}{3pt}
    \begin{tabular}{|p{1.8cm}|p{1.8cm}|p{1.8cm}|p{1.8cm}|p{1.8cm}|p{1.8cm}|p{1.8cm}|}
      \hline
      \multirow{2}{*}{Approach} & \multirow{2}{*}{\shortstack{Intelligent\\ Feedback}} &
      \multicolumn{3}{c|}{Critical  Points (\%)} &
      \multicolumn{2}{c|}{\shortstack{Improved\\ Performance (\%)}} \\
      \cline{3-5} \cline{6-7} &  & Yes & No & Overlap & Yes & No \\
      \hline
      RI & $3.33 \pm 1.23$ & $83.30 \pm 0.37$ & $16.70 \pm 0.37$ & $73.47 \pm 5.49$ &
      $91.70 \pm 0.28$ & $8.30 \pm 0.28$\\
      \hline
      AFS w/ \mbox{Random $\Omega$} & $2.82 \pm 0.94$ & $66.70 \pm 0.47$ & $33.30 \pm
      0.94$ & $75.51 \pm 5.27$ & $41.70 \pm 0.49$ & $58.30 \pm 0.49$\\
      \hline
      \mbox{AFS (Ours)} & $3.25 \pm 0.83$ & $100.00 \pm 0.00$ & $0.00 \pm 0.00$ & $81.63
      \pm 4.94$ & $100.00 \pm 0.00$ & $0.00\pm0.00$\\
      \hline
  \end{tabular}%
  \caption{\normalsize Participants' qualitative assessment from the pilot study on a simulated domain.}
  \label{tab:user_study_qual}
\end{table*}

This interface design allowed participants to provide both categorical and comparative feedback efficiently. After an initial training phase to practice providing feedback in different formats, participants self-reported the probability, $\psi(f)$, of providing feedback in a given format $f$, and effort ratings, $C(f)$.

\subsection{Results and Takeaways}
Fig.~\ref{fig:user_study_domain}(b) shows that our approach tends to result in fewer NSEs, compared to the baselines. Since the NSE penalty is an aggregate measure that obscures severity distribution, we report exact NSE encounters by category for this study.
Table~\ref{tab:user_study_qual} reports average over responses to our questions: ``On a scale of 1 to 5, how intelligent do you think the agent's choice of feedback formats are, given your preferences?'', ``Were the states in which the agent requested for feedback critical to its learning?'', and ``Did the agent's performance improve at the end of the learning phase?''. In addition, we also report the overlap between user-identified important query points and query points chosen by each approach. 

Overall, the results of this pilot study indicate that (1) AFS tends to effectively select query points and lead to improved learning outcomes, when operating under a learned feedback model; and (2) AFS's performance in this pilot study where users interact with a simulated agent is comparable to that of our results in simulation. Building on these results and the insights gained from this pilot study, we next conduct a user study where human participants interact with a physical robot. Such a setting will enable us to evaluate how well the observed trends extend to human-robot physical interactions, and how that affects the usability, trust and the users' perceived workload when interacting with a system that learns using AFS.

\section{In-Person User Study with Kinova Arm}
\label{sec:3}

\subsection{User Interface and Feedback Modalities}

\begin{figure*}[th]
    \centering
    \includegraphics[scale=0.7]{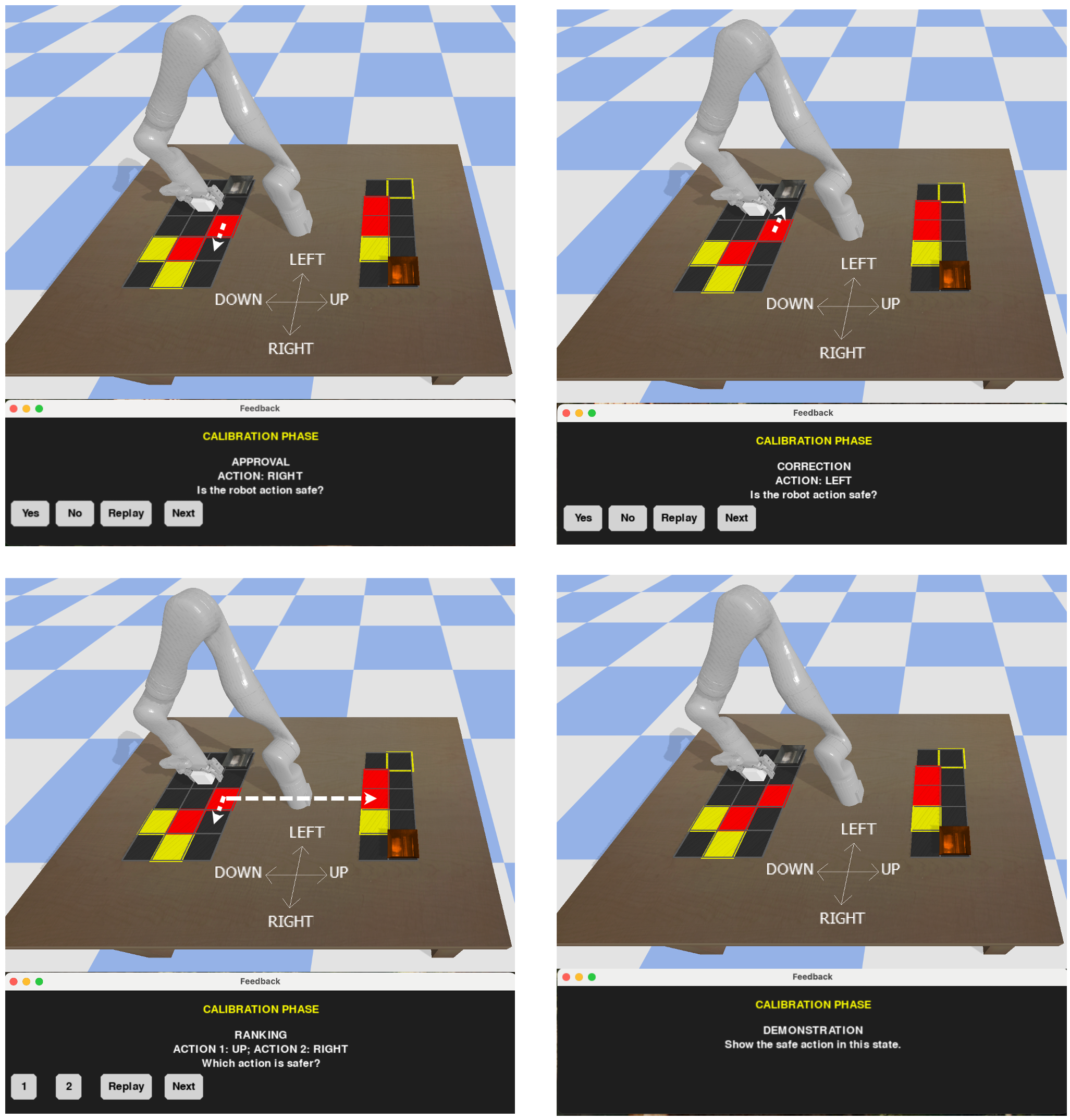}
    \caption{Interface and prompts used in the  user study with the Kinova arm. The interface displayed a short clip of the robot's action (indicated by the white dotted arrow) along with a dialog box prompting user feedback. \textbf{Top row:} Approval and Correction; \textbf{Bottom row:} Ranking and DAM formats.}
    \label{fig:kinova_interface}
    \vspace{-20pt}
\end{figure*}

Figure~\ref{fig:kinova_interface} illustrates the interface and feedback mechanisms used during the in-person user study with the Kinova Gen3 7-DoF robotic arm. Participants interacted with the robot through both a GUI and direct physical manipulation of the arm, depending on the feedback format being queried. 

\noindent \textbf{Interface Layout} The GUI displayed a simulation of the robot's motion on the tabletop, together with clearly labeled directional arrows (\textit{Up, Down, Left, Right}) corresponding to the robot's possible actions. The lower portion of the interface presented a dialog box, that indicates the current condition, the robot's selected action, and the question or instruction corresponding to the current format.
Each query consisted of a brief robot motion in simulation, followed by the corresponding GUI prompt. Participants could replay the motion before submitting their feedback.

\end{document}